\pdfoutput=1

\documentclass[11pt]{article}

\usepackage[final]{acl}

\usepackage{times}
\usepackage{latexsym}
\usepackage{amsmath}

\usepackage[T1]{fontenc}

\usepackage[utf8]{inputenc}

\usepackage{microtype}

\usepackage{inconsolata}
\usepackage{xcolor}
\usepackage{graphicx}
\usepackage{multirow}
\usepackage{booktabs}
\usepackage{rotating}
\usepackage{subcaption}
\usepackage{tcolorbox}
\usepackage{threeparttable}
\usepackage{pythonhighlight}
\usepackage{listings}
\definecolor{beaublue}{rgb}{1.0, 0.46, 0.44}
\definecolor{lightcyan}{rgb}{0.88, 1.0, 1.0}
\colorlet{mythmback}{lightcyan!40!white}
\newtcolorbox{boxEnv}{
colback=mythmback,coltitle=blue,colframe=mythmback,
center,
width=\linewidth,
boxrule=0.5pt,
left=5pt,right=0pt,
top=2pt,bottom=2pt,
before skip=10pt, after skip=10pt
}

\definecolor{codegreen}{rgb}{0,0.6,0}
\definecolor{codegray}{rgb}{0.5,0.5,0.5}
\definecolor{codepurple}{rgb}{0.58,0,0.82}
\definecolor{backcolour}{rgb}{0.95,0.95,0.92}

\title{AMR-Evol: Adaptive Modular Response Evolution Elicits Better Knowledge Distillation for Large Language Models in Code Generation}

\author{%
  $^{\spadesuit}$Ziyang Luo~,
  $^{\heartsuit}$Xin Li~,
  $^{\spadesuit}$Hongzhan Lin~\\[5pt]
  $^{\spadesuit}$Jing Ma\footnotemark[1],~
  $^{\heartsuit}$Lidong Bing\thanks{Corresponding Authors.}\\[5pt]
  $^\spadesuit$Hong Kong Baptist University, $^\heartsuit$Alibaba DAMO Academy\\[3pt]
  \texttt{\{cszyluo,majing\}@comp.hkbu.edu.hk ~~ binglidong@gmail.com}\\
}

\begin{document}
\maketitle

\begin{abstract}
    The impressive performance of proprietary LLMs like GPT4 in code generation has led to a trend to replicate these capabilities in open-source models through knowledge distillation (e.g. Code Evol-Instruct). However, these efforts often neglect the crucial aspect of response quality, relying heavily on teacher models for direct response distillation. This paradigm, especially for complex instructions, can degrade the quality of synthesized data, compromising the knowledge distillation process.
    To this end, our study introduces the \textbf{A}daptive \textbf{M}odular \textbf{R}esponse \textbf{Evol}ution (\textbf{AMR-Evol}) framework, which employs a two-stage process to refine response distillation. The first stage, modular decomposition, breaks down the direct response into more manageable sub-modules. The second stage, adaptive response evolution, automatically evolves the response with the related function modules.
    Our experiments with three popular code benchmarks—HumanEval, MBPP, and EvalPlus—attests to the superiority of the \textbf{AMR-Evol} framework over baseline response distillation methods. By comparing with the open-source Code LLMs trained on a similar scale of data, we observed performance enhancements: more than +3.0 points on HumanEval-Plus and +1.0 points on MBPP-Plus, which underscores the effectiveness of our framework. Our codes are available at \url{https://github.com/ChiYeungLaw/AMR-Evol}.
\end{abstract}
\begin{figure}
    \centering
    \includegraphics[width=0.48\textwidth]{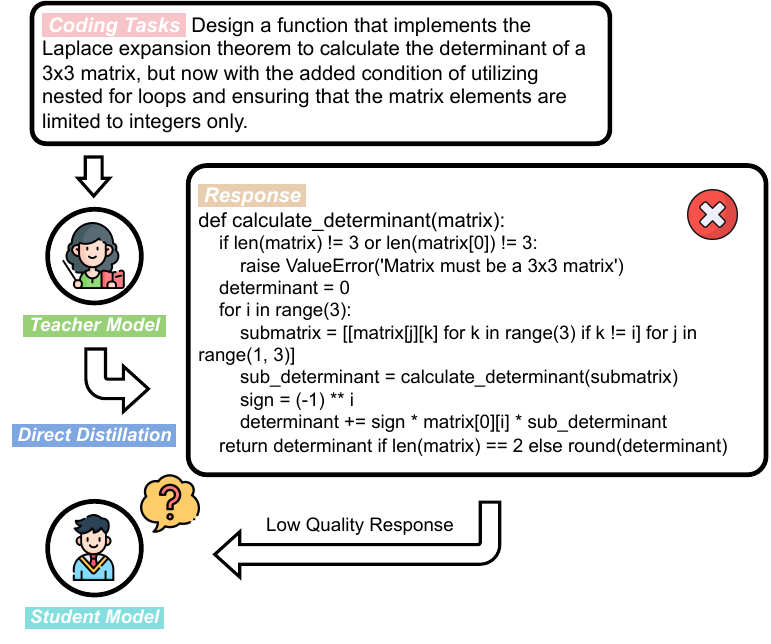}
    \caption{Direct distillation from the teacher model possibly yields low quality responses for complex tasks, thereby causing confusion within the student model.}
    \label{fig:intro}
\end{figure}

\section{Introduction}\label{sec:intro}
Recently, the powerful proprietary large language models (LLMs), like GPT3~\cite{gpt3}, GPT4~\cite{GPT4}, Gemini~\cite{gemini} and Claude~\cite{claude}, have showcased impressive code generation ability. Especially, GPT4, the most performant model, has recorded pass rates exceeding 85\% on the well-known HumanEval benchmark~\cite{humeval}. Despite their strengths, the closed-source nature sparks accessibility and privacy concerns~\cite{gpt_privacy}. In response, there is a trend of adopting knowledge distillation~\cite{KD_LLM} to transfer the advanced code generation ability from the proprietary LLMs to open-source counterparts, thereby enhancing their capabilities while ensuring broader availability and owner autonomy.

Given that accessing the model weights of proprietary LLMs is infeasible, the knowledge distillation pipeline is considered as a process where the teacher models synthesize supervised data, primarily consisting of instruction-response pairs~\cite{data_synthetic}. Student models are subsequently trained on this data, enabling the transfer of capabilities from the teacher models. For example, \citet{codealpaca} employs the self-instruct method~\cite{wang2022self} to prompt the teacher model to generate new coding instructions based on predefined seed tasks. Similarly, OSS-Instruct~\cite{magiccoder} utilizes a variety of code snippets sourced from GitHub to inspire GPT-3.5 to produce novel coding instructions. Likewise, Code Evol-Instruct~\cite{wizardcoder} employs iterative prompting to progressively elevate the complexity of code instructions provided by teacher models. Each of these methods has proven effective in distilling coding knowledge from teacher models.

Despite these advancements, there remains an unresolved challenge in enhancing the quality of code response distillation within the data synthesis process. In this setting, code responses serve as labels that teach the student models. Previous works have shown that higher-quality responses can lead to more effective distillation~\cite{LIMA,ORCA}. However, current methods~\cite{codealpaca,magiccoder,wizardcoder} tend to rely solely on teacher models for direct response distillation. As shown in Figure~\ref{fig:intro}, this approach is limited by the capabilities of the teacher models, making it difficult to produce accurate responses for complex tasks. The issue becomes even more challenging with methods like Code Evol-Instruct, which deliberately amplify the complexity of instructions. Consequently, relying on direct distillation can result in lower-quality responses, ultimately affecting the performance of the student models~\cite{DBLP:journals/corr/abs-2402-05123}.

A straightforward yet costly solution to guarantee response quality is to hire human annotators to craft the unit tests for each response. These tests could then be used in an execution-based strategy to validate answers. However, this method is financially prohibitive because it requires the recruitment of annotators with extensive programming expertise. Alternatively, depending on the teacher model to automatically generate unit tests for self-repair~\cite{CodeT,self-repair,DBLP:journals/corr/abs-2304-05128} introduces the same concern of response quality, providing no certainty regarding the correctness of the code repair.

To address the challenge of distilling high-quality code responses from teacher models, we introduce a novel framework named \textbf{A}daptive \textbf{M}odular \textbf{R}esponse \textbf{Evol}ution (\textbf{AMR-Evol}). In Figure~\ref{fig:intro}, the example reveals that the direct response distillation can somewhat capture the essential concepts required for solving coding tasks; however, it often deviates from the specific requirements and incorporates logical errors. Motivated by this observation, \textbf{AMR-Evol} leverages the outputs of direct distillation as seed data and employs a two-stage process—namely, modular decomposition and adaptive response evolution—to gradually refine the distilled code responses. By intricately refining the process of response distillation, our framework elicits better knowledge distillation of the student models.

In the first stage of our \textbf{AMR-Evol}, we adopt the idea from modular programming~\cite{DBLP:conf/sosp/Dijkstra67} to manage the complexity of distilling code responses. By utilizing direct responses as the seeds, this method breaks down the coding task into smaller, more manageable sub-modules. This strategy shifts the focus of the teacher models towards solving these sub-modules step-by-step rather than generates a complete solution in a single attempt, whose effectiveness has been verified in recent Chain-of-X works~\cite{cot,CodeChain,Chain-of-X}.

Additionally, while coding tasks may vary significantly in objectives, the modular components need to construct their solutions frequently exhibit commonalities, or can even be identical~\cite{DBLP:journals/cacm/Parnas72a}. Hence, our adaptive response evolution stage leverages an auxiliary functional module database to store all validated modules for reuse. During response generation, this process utilizes the modules formulated in the decomposition stage to retrieve suitable, pre-validated modules from the database. These related modules serve as in-context examples, aiding the adaptive refinement of responses, thus reducing our sole reliance on teacher models. As evolution progresses, any newly created modules that differ from those in the database are added after a verification process by the teacher model.

We apply our \textbf{AMR-Evol} framework to different student models and select the most representative coding benchmarks, including HumanEval~\citep{humeval}, MBPP~\cite{MBPP}, and EvalPlus~\cite{humanevalp}, for evaluation. The results reveal that our \textbf{AMR-Evol} framework consistently surpasses other response distillation methods, namely direct response distillation, chain-of-thought distillation, and response repairing. These results affirm the superiority of our approach in improving knowledge distillation for LLMs in code generation. Moreover, by integrating our \textbf{AMR-Evol} with Code Evol-Instruct, one of the SOTA instruction construction methods, our models achieve better performance than the open-source alternatives trained on a comparable data scale. Specifically, we observed an improvement of more than +3.0 on HumanEval-Plus and +1.0 on MBPP-Plus.
\section{Related Work}

\paragraph{LLMs and Code Generation.} 

Recently, LLMs have showcased significant achievements across a vast array of tasks. Leading tech firms have made substantial progress in developing highly advanced close-source LLMs, including OpenAI's GPT4~\cite{GPT4}, Google's PaLM~\cite{PaLM,palm2} and Gemini~\cite{gemini}, as well as Anthropic's Claude~\cite{claude}. On the other side, the AI community has also seen the launch of several open-source LLMs, with model weights becoming publicly available. MistralAI has contributed the Mistral-Series~\cite{mistral}. Google has released UL2-20B~\cite{UL2} and Gemma~\cite{gemma}. Tsinghua University introduced GLM-130B~\cite{GLM-130B} and MiniCPM~\cite{MiniCPM}, while Meta has made available OPT~\cite{opt} and LLaMA1\&2\&3~\cite{llama,llama2,llama3}. Furthermore, Allen AI has introduced the wholly open-sourced LLM, OLMo~\cite{OLMO}, and Microsoft has released Phi-series~\cite{phi1,DBLP:journals/corr/abs-2309-05463}. Although a gap remains between the open-source models and their closed-source counterparts, this gap is gradually narrowing.

In parallel, recent research efforts have been directed towards leveraging LLMs for code-related tasks to address the understanding and generation of code. OpenAI has unveiled Codex~\citep{humeval}, Google has proposed CodeGemma~\cite{codegemma}, and Salesforce has introduced CodeGen-Series~\cite{codegen,codegen2}, and CodeT5\&Plus~\cite{codet5,CodeT5+}. Contributions from Tsinghua University include CodeGeeX~\cite{CodeGeeX}, and the BigCode Project has developed StarCoder1\&2~\cite{li2023starcoder,starcoder2}. Meta has also made its mark with the CodeLlama~\cite{codellama}, while DeepSeek has open-sourced the DeepSeekCoder~\cite{deepseekcoder}. These initiatives underscore the growing interest in employing powerful base LLMs for code generation. Our work introduces a novel method for more effectively distilling code knowledge from closed-source models to these open-source base models, thereby enhancing the coding performance.

\paragraph{Knowledge Distillation for Code Generation.} To enhance the capabilities of open-source LLMs for code generation, recent works have adopted the knowledge distillation paradigm, utilizing closed-source LLMs as teachers for supervised data synthesis~\cite{DBLP:journals/corr/abs-2310-18628,DBLP:journals/corr/abs-2402-14658,li2024instructcoder,yuan2024advancing}. For example, \citet{codealpaca} employs the self-instruct method~\cite{wang2022self} to generate training data, while Magicoder~\cite{magiccoder} generates training content using code snippets from GitHub. WizardCoder~\cite{wizardcoder}, on another hand, introduces the Code Evol-Instruct approach to progressively increase the complexity of coding tasks. Despite these advancements, a common limitation among these efforts is their primary focus on the creation of code instructions, often overlooking the criticality of enhancing code response distillation. Our research takes an orthogonal path by concentrating on the refinement of code response distillation, offering a novel perspective compared to previous works.
\begin{figure*}
    \centering
    \includegraphics[width=\textwidth]{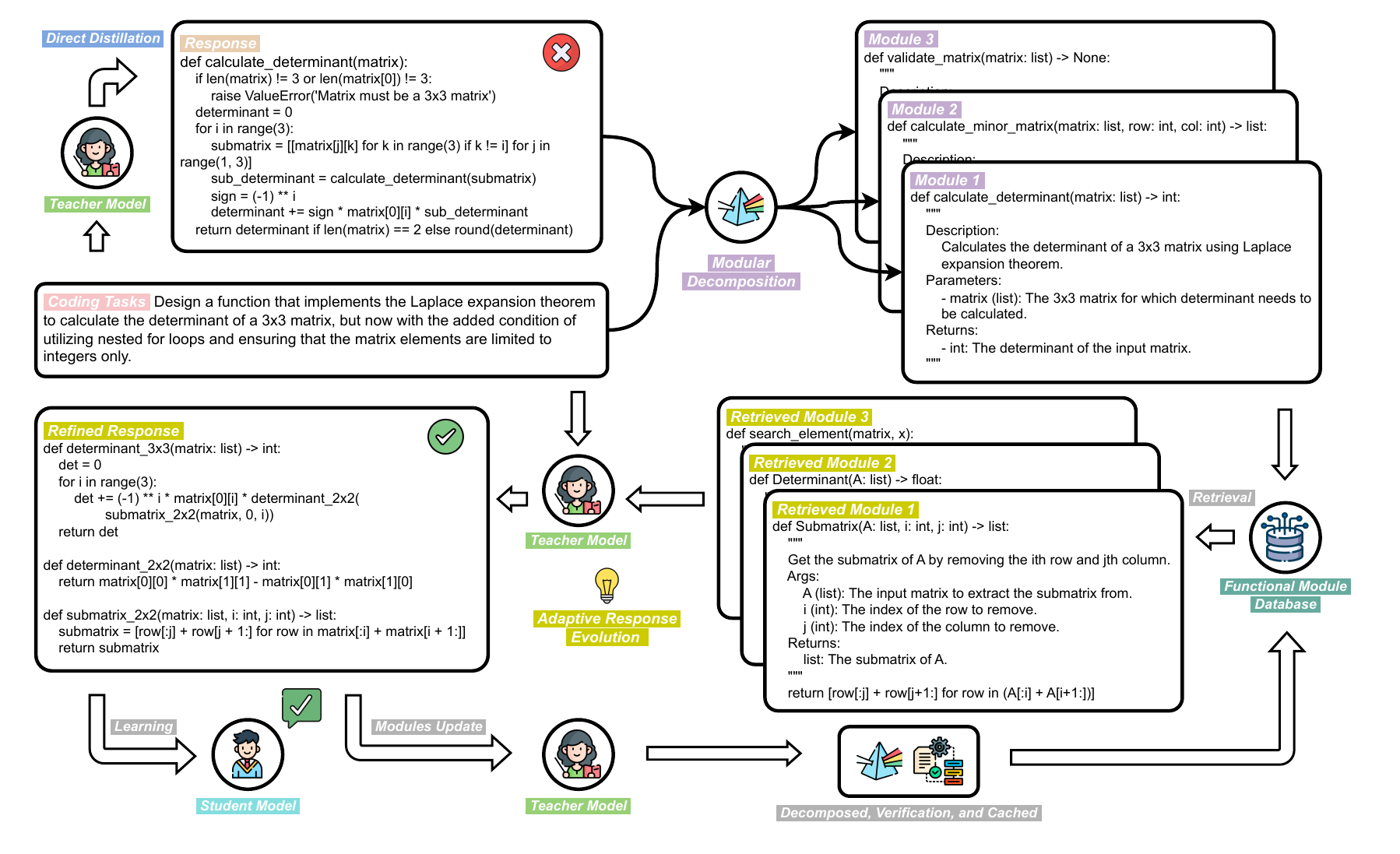}
    \caption{Our \textbf{Adaptive Modular Response Evolution (AMR-Evol)} framework with modular decomposition and adaptive response evolution elicits better response distillation for LLMs in code generation.}
    \label{fig:method}
\end{figure*}

\section{Method}

As depicted in Figure~\ref{fig:method}, we introduce our novel framework, \textbf{AMR-Evol}, aimed at improving code response distillation to elicit better performance of the student models. In this section, we will provide a detailed discussion of our framework's pipeline.

\subsection{Direct Response Distillation}\label{sec:direct}

In the knowledge distillation framework, the foremost goal is enabling the student model $\mathcal{M}_{s}$ to assimilate the strategies deployed by the teacher model $\mathcal{M}_{t}$ in tackling code generation tasks. Utilizing approaches like Code Evol-Instruct facilitates the generation of an extensive dataset of code instructions $\{I\}$ by the teacher model. Subsequently, the direct response distillation method employs the teacher model to process these task instructions to produce the corresponding code responses $R_d$, resulting in a paired dataset, $\mathcal{D}_{direct}=\{(I, R_d)\}$. Then, the student model $\mathcal{M}_{s}$ learns from this dataset through supervised fine-tuning.

\subsection{Adaptive Modular Response Evolution}

As discussed in Section~\ref{sec:intro}, direct responses $\mathcal{D}_{direct}$ to complex instructions can result in suboptimal quality, which in turn impacts the performance of the student model $\mathcal{M}_{s}$. While these responses often include logical errors or may not fully align with the precise requirements of the tasks, they generally remain close to correct and capture the essential concepts needed for task solution. To address this, our \textbf{AMR-Evol} framework capitalizes on these direct response distillations as a starting point. It incorporates a two-stage method—modular decomposition and adaptive response evolution—for an automated refinement process that improves the quality of responses, thereby enhancing the efficacy of distillation.

\paragraph{Modular Decomposition (MD).} In the first stage of our framework, we employ the principle of modular programming~\cite{DBLP:conf/sosp/Dijkstra67} to tackle the complexity inherent in distilling code responses. Our method utilizes direct responses $R_d$ as a starting point, guiding the teacher model $\mathcal{M}_t$ in breaking down the given code instructions into a series of smaller, well-defined sub-modular functions. We represent this process mathematically as follows:
\begin{equation}
    \{F_1^m, F_2^m, \ldots, F_n^m\} \leftarrow \mathcal{M}_t\left(I, R_d\right),
\end{equation}
where each function module $F_i^m$ is conceptualized to fulfill a distinct subset of requirements stipulated by the code instruction $I$. This decomposition breaks down complex instructions into a series of easier and more manageable sub-modules, enabling the teacher model to tackle each one with less difficulty. This results in a more effective response distillation process.

\paragraph{Adaptive Response Evolution (ARE).} In the second stage, we observe that while coding instructions may greatly differ, the sub-modules needed for assembling the final solution often share similarities or can even be identical~\cite{DBLP:journals/cacm/Parnas72a}. Leveraging this insight, we establish an auxiliary functional module database $\{F_i^v\}$, which archives all validated modules for future reuse. This database acts as a repository, enabling the retrieval of previously validated sub-modules to foster the creation of new code responses.

Building upon the modular decomposition achieved in the first stage, $\{F_1^m, F_2^m, \ldots, F_n^m\}$, we initially convert both the newly decomposed and previously archived functional modules into dense vector representations through a sentence embeddings model $\mathcal{M}_r$:
\begin{equation}
    V_{f_i^{(\cdot)}} \leftarrow \mathcal{M}_r\left(F_i^{(\cdot)}\right),
\end{equation}
where $V_{f_i^{(\cdot)}}$ denotes the dense representation of any given functional module $F_i^{(\cdot)}$. Then, to facilitate the retrieval of the most suitable archived module for each new sub-module, we apply:
\begin{equation}
    \text{Sim}\left(F_i^m, F_j^{v}\right) \leftarrow \text{CosineSimilarity}\left(V_{f_i^m}, V_{f_j^v}\right),
\end{equation}
where $\text{Sim}\left(F_i^m, F_j^{v}\right)$ calculates the similarity between the dense representations of two modules using cosine similarity. The archived modules that exhibit the highest similarity are then used as additional in-context contents, assisting the teacher model in refining the final code responses:
\begin{equation}
    R_{amr} \leftarrow \mathcal{M}_t\left(I, \{F_i^m\}, \{F_i^v\}\right),
\end{equation}
where $R_{amr}$ represents the refined code responses. These responses, alongside the original instruction $I$, compile an evolved dataset aimed at optimizing the knowledge distillation process.

As the process evolves, our framework identifies new modules within $R_{amr}$ that exhibit notable differences from those currently in the database—judged by the cosine similarity between the new modules and existing ones. Modules that are distinct undergo a rigorous verification stage prior to their integration into the database. This critical stage harnesses the capabilities of the teacher model for generating unit tests tailored to the functionalities of the specific modules. This procedure not only assesses the functional correctness of the new modules but also ensures that they meet the predefined quality standards, thereby streamlining the process of enriching the module database with reliable and effective components.

\paragraph{Functional Module Database.} The functional module database is pivotal within our \textbf{AMR-Evol} framework. We begin by compiling a collection of seed functions that have been validated. Leveraging the self-instruct method~\cite{wang2022self}, we prompt our teacher models to generate a diverse range of function modules. Following this, we adopt a strategy similar to CodeT~\cite{CodeT}, instructing the teacher models to produce unit tests that verify the functionality of these modules. Only the functions that pass these unit tests are included in our dataset. Through this stringent process, we construct a seed functional module database that becomes a fundamental component of our framework.

\subsection{Knowledge Distillation}

Upon completing the data synthesis process with the help of teacher models, we acquire a dataset that consists of paired instructions and responses, $\mathcal{D}_{amr}=\{(I,R_{amr})\}$. This dataset equips the student model $\mathcal{M}_{s}$ for the task of knowledge distillation, where it is trained to use $I$ as input with the goal of generating responses $R_{amr}$ that closely resemble those produced by the teacher model. The training follows an auto-regressive learning objective, formalized as follows:
\begin{equation}
    \mathcal{L}(\theta) = -\sum_{(I,R_{amr}) \in \mathcal{D}_{amr}} \log P(R_{amr} | I; \theta),
\end{equation}
where $\mathcal{L}(\theta)$ denotes the loss function minimized during training, and $\theta$ signifies the parameters of the student model $\mathcal{M}_{s}$. This objective encourages the student model to accurately predict the next token in the response sequence, given the instruction $I$ and the current state of the generated response.
\section{Experiment}

\subsection{Setup}

\paragraph{Baselines.} Within our evaluation framework, we compare the performance of our framework against several baselines in code response distillation. The first of these, referred to as \textbf{direct}, utilizes teacher models to distill code responses in a straightforward manner, as detailed in Section~\ref{sec:direct}. The second baseline employs the Chain-of-Thought (\textbf{CoT}) prompting method for distilling responses~\cite{DBLP:conf/acl/HsiehLYNFRKLP23}. This approach is analogous to the few-shot CoT method~\cite{cot}, in which the teacher model first provides a step-by-step explanation leading up to the formulated response. Our third baseline, \textbf{AnsRepair}, draws inspiration from previous works~\cite{CodeT,self-repair,self-debug}, where the teacher models are utilized to generate unit tests. These tests serve to evaluate the correctness of the generated responses. If the responses fail these tests, the teacher models are subsequently invoked to make the necessary corrections. More details about baseline methods are included in the Appendix~\ref{app:baseline}.

\begin{table}
    \small
    \centering
    \begin{tabular}{@{}lcccc@{}}
        \toprule
        \textbf{Method} & \textbf{HE} & \textbf{HE-Plus} & \textbf{MBPP} & \textbf{MBPP-Plus}\\
        \midrule
        \midrule
        \multicolumn{5}{@{}l}{\textit{Complexity Level 1}}\\
        \textbf{Direct} & 54.9 & 46.3 & 65.9 & 54.1\\
        \textbf{CoT} & 52.4 & 45.7 & 65.7 & 53.4\\
        \textbf{AnsRepair} & 53.7 & 45.1 & 63.2 & 52.1\\
        \midrule
        \textbf{AMR-Evol} & 58.5 & 49.4 & 68.7 & 58.1\\
        ~$\Delta$ & +3.6 & +3.1 & +2.8 & +4.0\\
        \midrule
        \multicolumn{5}{@{}l}{\textit{Complexity Level 2}}\\
        \textbf{Direct} & 53.7 & 46.3 & 64.4 & 52.6\\
        \textbf{CoT} & 54.9	& 46.3 & 65.7 & 53.9\\
        \textbf{AnsRepair} & 56.1 & 47.6 & 63.4 & 52.9\\
        \midrule
        \textbf{AMR-Evol} & 56.1 & 47.6 & 68.7 & 56.6\\
        ~$\Delta$ & +0.0 & +0.0 & +3.0 & +2.7\\
        \midrule
        \multicolumn{5}{@{}l}{\textit{Complexity Level 3}}\\
        \textbf{Direct} & 52.4 & 45.7 & 65.2 & 53.9\\
        \textbf{CoT} & 52.4	& 43.9 & 65.7 & 53.9\\
        \textbf{AnsRepair} & 55.5 & 47.6 & 65.4 & 53.1\\
        \midrule
        \textbf{AMR-Evol} & 56.1 & 49.4 & 67.7 & 56.4\\
        ~$\Delta$ & +0.6 & +1.8 & +2.0 & +2.5\\
        \bottomrule
    \end{tabular}
    \caption{Comparison of various response distillation methods for code generation, utilizing \texttt{deepseek-coder-6.7b-base} as the student model.}
    \label{tab:ds_compare}
\end{table}

\begin{table}[h!]
    \small
    \centering
    \begin{tabular}{@{}lcccc@{}}
        \toprule
        \textbf{Method} & \textbf{HE} & \textbf{HE-Plus} & \textbf{MBPP} & \textbf{MBPP-Plus}\\
        \midrule
        \midrule
        \multicolumn{5}{@{}l}{\textit{Complexity Level 1}}\\
        \textbf{Direct} & 36.6 & 31.1 & 54.4 & 44.1\\
        \textbf{CoT} & 36.0 & 31.1 & 55.1 & 45.6\\
        \textbf{AnsRepair} & 35.4 & 29.3 & 56.4 & 45.4\\
        \midrule
        \textbf{AMR-Evol} & 37.8 & 32.3 & 57.4 & 45.6\\
        ~$\Delta$ & +1.2 & +1.2 & +1.0 & +0.0\\
        \midrule
        \multicolumn{5}{@{}l}{\textit{Complexity Level 2}}\\
        \textbf{Direct} & 37.2 & 31.1 & 55.4 & 44.6\\
        \textbf{CoT} & 36.0 & 31.1 & 54.6 & 45.6\\
        \textbf{AnsRepair} & 35.4 & 29.3 & 56.6 & 45.9\\
        \midrule
        \textbf{AMR-Evol} & 39.6 & 32.3 & 59.4 & 47.6\\
        ~$\Delta$ & +2.4 & +1.2 & +2.8 & +1.7\\
        \midrule
        \multicolumn{5}{@{}l}{\textit{Complexity Level 3}}\\
        \textbf{Direct} & 36.0 & 30.5 & 56.4 & 45.6\\
        \textbf{CoT} & 37.2 & 30.5 & 55.6 & 46.4\\
        \textbf{AnsRepair} & 37.2 & 29.3 & 55.6 & 44.9\\
        \midrule
        \textbf{AMR-Evol} & 39.0 & 32.9 & 59.1 & 46.9\\
        ~$\Delta$ & +1.8 & +2.4 & +2.7 & +0.5\\
        \bottomrule
    \end{tabular}
    \caption{Comparison of various response distillation methods for code generation, utilizing \texttt{CodeLlama-7b-hf} as the student model.}
    \label{tab:cl_compare}
\end{table}

\paragraph{Datasets and Benchmarks.} Our framework focuses on distilling responses and necessitates a dataset of instructions. To this end, we utilize a subset of the training set from the MBPP as our seed data. This is then expanded using the self-instruct method with the teacher model to generate around 10k instructions. With these newly derived instructions, we employ a process akin to the Code Evol-Instruct to iteratively synthesize a spectrum of complex coding instructions across three distinct levels of complexity. This variety allows us to assess our framework's efficacy in handling complex instructions. More data construction and decontamination details can be found in the Appendix~\ref{app:datasets}.

For performance evaluation, we utilize the well-known coding benchmark, namely HumanEval~\cite{humeval}, MBPP~\cite{MBPP}, and EvalPlus~\cite{humanevalp}. HumanEval contains 164 coding problems with an average of 9.6 test cases per problem. MBPP includes 399 coding problems, each with three automated test cases. EvalPlus extends the number of test cases for both HumanEval and MBPP, resulting in enhanced versions named HumanEval-Plus and MBPP-Plus. Following EvalPlus, we report our method's effectiveness in terms of pass rates using greedy decoding, which helps minimize the impact of any randomness in the results. More details are included in the Appendix~\ref{app:benchmark}.

\paragraph{Implementation Details.} For all experiments, we employ OpenAI's close-sourced LLM, \texttt{gpt-3.5-turbo-1106} as our teacher model and choose two popular open-sourced code LLMs, \texttt{deepseek-ai/deepseek-coder-6.7b-base} \cite{deepseekcoder} and \texttt{meta-llama/CodeLlama-7b-hf} \cite{codellama} as our student models. For the dense embeddings, we adopt one of the SOTA embeddings models, \texttt{Alibaba-NLP/gte-large-en-v1.5}~\cite{li2023towards} as our representation model. The supervised knowledge distillation phases of all experiments are conducted with 200 training steps, 3 epochs, a sequence length of 2048 and the AdamW optimizer~\cite{adamw}. For further training details and prompting designes, please refer to the Appendix~\ref{app:impl}.

\begin{figure*}[h]
    \centering
    \begin{subfigure}[b]{0.3\textwidth}
        \includegraphics[width=\textwidth]{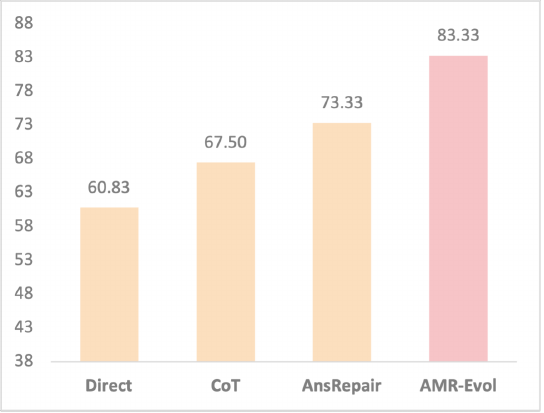}
        \caption{Complex 1.}
        \label{fig:image1}
    \end{subfigure}
    \hfill 
    \begin{subfigure}[b]{0.3\textwidth}
        \includegraphics[width=\textwidth]{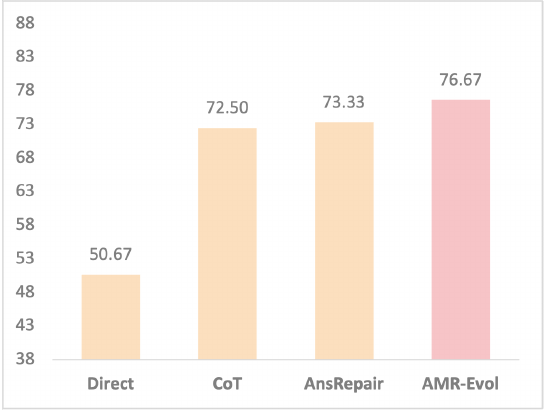}
        \caption{Complex 2.}
        \label{fig:image2}
    \end{subfigure}
    \hfill 
    \begin{subfigure}[b]{0.3\textwidth}
        \includegraphics[width=\textwidth]{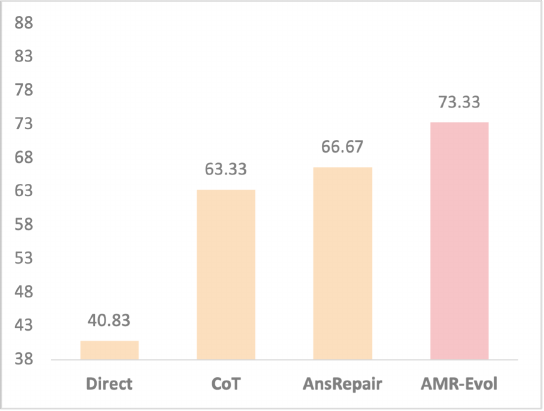}
        \caption{Complex 3.}
        \label{fig:image3}
    \end{subfigure}
    \caption{Manual evaluation of the accuracy of various code response distillation methods across 120 randomly selected samples from each complexity level.}
    \label{fig:human}
\end{figure*}

\subsection{Main Results}

In Table~\ref{tab:ds_compare}, our \textbf{AMR-Evol} consistently outperforms various response distillation methods for code generation, when adopt the \texttt{deepseek-coder-6.7b-base} as the student model. Specifically, at Complexity Level 1, \textbf{AMR-Evol} exhibited superior results, with improvements ranging between +2.8 to +4.0 across all tasks. Our method maintained this lead in Complexity Level 2, with the most substantial gains in MBPP and MBPP-Plus, at +3.0 and +2.7, respectively. Notably, even at the highest complexity (Level 3), the method continued to show incremental enhancements, most prominently a +2.5 increase in MBPP-Plus. The performance exhibits \textbf{AMR-Evol's} consistent proficiency in eliciting better code knowledge distillation across varying degrees of complexity.

When utilizing \texttt{CodeLlama-7b-hf} as the student model, Table~\ref{tab:cl_compare} reveals that the performance patterns of \textbf{AMR-Evol} closely paralleled its efficacy with the previous model. Albeit with modest improvements at Complexity Level 1, \textbf{AMR-Evol} showed more enhancement in higher complexity scenarios. At Complexity Level 2, our method achieves increases of +2.4 on HE and +2.8 on MBPP. The upward trend persisted through Complexity Level 3, as the method underscored its robustness with increases such as +2.4 on HE-Plus and +2.7 on MBPP. These results solidify \textbf{AMR-Evol} as an effective method for code knowledge distillation, adaptable to various instruction complexity levels.

\subsection{Analysis}

\paragraph{Quality Comparison.} Our experimental findings illustrate the effectiveness of our \textbf{AMR-Evol} in enhancing the knowledge distillation. To further validate the efficacy of \textbf{AMR-Evol} in producing better instruction fine-tuning data, we conducted a manual evaluation. We randomly selected the sample sets of 120 coding problems for each levels of complexity. Given that all samples are coding challenges, their responses can be definitively classified as either correct or incorrect. Two experienced programmers were engaged to review and label the code responses generated by various methods as suitable or not. The manual assessment results, depicted in Figure~\ref{fig:human}, reveal that although no method attained complete perfect, \textbf{AMR-Evol} demonstrated consistently superior performance compared to all other baseline methods across all complexity levels. In Appendix~\ref{app:qualitative}, we also include some examples of responses generated by different methods to qualitatively compare their quality.

\begin{table}
    \small
    \centering
    \begin{tabular}{@{}lcccc@{}}
        \toprule
        \textbf{Method} & \textbf{HE} & \textbf{HE-Plus} & \textbf{MBPP} & \textbf{MBPP-Plus}\\
        \midrule
        \midrule
        \multicolumn{5}{@{}l}{\textit{Complexity Level 1}}\\
        \textbf{AMR-Evol} & 58.5 & 49.4 & 68.7 & 58.1\\
        ~~w/o MD & 57.9 & 49.4 & 67.4 & 55.9\\
        ~~w/o ARE & 56.1 & 48.8 & 69.4 & 57.1\\
        \midrule
        \multicolumn{5}{@{}l}{\textit{Complexity Level 2}}\\
        \textbf{AMR-Evol} & 56.1 & 47.6 & 68.7 & 56.6\\
        ~~w/o MD & 54.9 & 46.3 & 67.7 & 54.4\\
        ~~w/o ARE & 54.9 & 47.0 & 67.4 & 55.9\\
        \midrule
        \multicolumn{5}{@{}l}{\textit{Complexity Level 3}}\\
        \textbf{AMR-Evol} & 56.1 & 49.4 & 67.7 & 56.4\\
        ~~w/o MD & 54.3 & 47.6 & 66.4 & 53.6\\
        ~~w/o ARE & 53.0 & 47.0 & 67.4 & 54.6\\
        \bottomrule
    \end{tabular}
    \caption{Ablation studies by removing modular decomposition (MD) or adaptive response evolution (ARE) in our framework.}
    \label{tab:ablation}
\end{table}

\paragraph{Ablation.} In Table~\ref{tab:ablation}, we present an ablation study meticulously designed to identify the individual contributions of modular decomposition (MD) and adaptive response evolution (ARE) to the efficacy of our framework. First, we remove the MD stage in our framework by adopting direct response to retrieve the related function modules for ARE. This led to a performance drop, underscoring its crucial role in our framework. Specifically, the omission of MD typically results in the recall of only one function module based on the direct response. However, while direct responses address more complex or larger coding tasks, function modules target tasks with finer granularity. This difference creates a gap, making it challenging for the retrieved function modules to effectively contribute to refining the direct responses.

\begin{table*}
    \centering
    \begin{tabular}{lcccccc}
        \toprule
        \textbf{Model} & \textbf{Size} & \textbf{\#SFT Ins} & \textbf{HE} & \textbf{HE-Plus} & \textbf{MBPP} & \textbf{MBPP-Plus}\\
        \midrule
        \midrule
        \multicolumn{7}{l}{\textit{Proprietary models}}\\
        \textbf{GPT4} & - & - & 85.4 & 81.7 & 83.0 & 70.7\\
        \textbf{GPT3.5} & - & - & 72.6 & 65.9 & 81.7 & 69.4\\
        \textbf{Gemini Pro} & - & - & 63.4 & 55.5 & 72.9 & 57.9\\
        \midrule
        \multicolumn{7}{l}{\textit{Base model: deepseek-ai/deepseek-coder-6.7b-base}}\\
        $\dagger$\textbf{DeepSeekCoder-Instruct} & 6.7B & >1M & 73.8 & 70.1 & 72.7 & 63.4\\
        \textbf{MagiCoder-DS} & 6.7B & 75k & 63.4 & 57.3 & 75.2 & 61.9\\
        $\ddagger$\textbf{WaveCoder-DS} & 6.7B & 20k & 66.5 & 57.9 & 73.7 & 60.4\\
        \midrule
        \textbf{DeepSeekCoder-AMR-Evol} & 6.7B & 50k & 68.9 & 61.0 & 74.4 & 62.9\\
        \midrule
        \multicolumn{7}{l}{\textit{Base model: meta-llama/CodeLlama-7b-Python-hf}}\\
        $\dagger$\textbf{CodeLlama-Instruct} & 7B & 80k & 32.9 & 26.8 & 59.1 & 45.6\\
        \textbf{WizardCoder-CL} & 7B & 78k & 55.5 & 48.2 & 64.9 & 53.9\\
        \textbf{MagiCoder-CL} & 7B & 75k & 54.3 & 48.8 & 63.7 & 51.9\\
        \midrule
        \textbf{CodeLlama-AMR-Evol} & 7B & 50k & 59.1 & 51.8 & 64.7 & 55.4\\
        \bottomrule
        \multicolumn{7}{l}{$\dagger$: Official instruction models. Responses are distilled from unknown, humans or themselves.}\\
        \multicolumn{7}{l}{$\ddagger$: Responses are distilled from GPT4.}\\
    \end{tabular}
    \caption{Comparison of our fine-tuned models against both publicly available academic Code LLMs, similarly scaled in terms of SFT data and based on the same student models as ours, and the official instruction-based LLMs. We either download the model weights or utilize the APIs for performance reproduction.}
    \label{tab:compare_oss}
\end{table*}
Subsequently, we exclude the ARE stage, which also resulted in a performance decline, highlighting its vital role in the framework. Without ARE, the generation of responses is solely reliant on the modular decomposition output, lacking the improvements that come from in-context learning with related function modules. This places the entire responsibility for refining responses on the inherent capabilities of the teacher model. This analysis strongly reinforces the indispensable nature of both MD and ARE within our framework. In Appendix~\ref{app:mod_example}, we also present examples to showcase the output of the MD stage and the top-1 function modules retrieved from the database.

\subsection{Comparing with Open Code LLMs}

To delve deeper into the efficacy of our framework, we have incorporated \textbf{AMR-Evol} with one of the SOTA instruction construction methods, Code Evol-Instruct, to expand our SFT data set. We have generated around 50k instructions using this approach and employed \textbf{AMR-Evol} to distill code responses from the teacher models (GPT3.5). Subsequently, we used \texttt{deepseek-coder-6.7b-base} and \texttt{CodeLlama-7b-Python-hf} as our two student models for training. For a relative fair comparison, we compare our fine-tuned student models against publicly available academic Code LLMs, which are trained with a similar scale of SFT data and employ the same base models as ours. This includes MagiCoder-DS/CL~\cite{magiccoder}, WaveCoder-DS~\cite{wavecoder}, and WizardCoder-CL~\cite{wizardcoder}. We also compare against official instruction models, namely DeepSeek-Coder-Instruct and CodeLlama-Instruct, to showcase performance gaps. For more discussions about baseline selection and SFT details, please refer to the Appendix~\ref{app:open}.

Table \ref{tab:compare_oss} showcases the exceptional performance of DeepSeekCoder-AMR-Evol across all tasks. When compared to MagiCoder-DS, trained with 75k SFT data, and WaveCoder-DS, distilled from GPT4, the AMR-Evol version notably stands out by demonstrating substantial performance gains: +2.4 on HE, +3.2 on HE-Plus, and +1.0 on MBPP-Plus. Even when compared to the official instruction model, which is trained with more than 20 times as much data, our model achieves comparable performance on MBPP and MBPP-Plus. Similarly, the CodeLlama-AMR-Evol variant exhibits superior performance in most tasks, with performance improvements of +3.6 on HE, +3.0 on HE-Plus, and +1.5 on MBPP-Plus, respectively. Moreover, our model significantly outperforms CodeLlama-Instruct, which is an official model from Meta. In addition, the Pass@k sampling results, presented in Appendix~\ref{app:open},  Table~\ref{tab:compare_sampling}, also evident the better performance of our models.

\begin{table}
    \small
    \centering
    \begin{tabular}{lccc}
        \toprule
        \textbf{Model} & \textbf{CC Val} & \textbf{CC Test} & \textbf{APPS}\\
        \midrule
        \midrule
        \textbf{DS-Instruct} & 7.69 & 6.67 & 11.67\\
        \textbf{MagiCoder-DS} & 8.55 & 12.73 & 13.00\\
        \textbf{DS-AMR-Evol} & 10.26 & 12.73 & 14.22\\
        \bottomrule
    \end{tabular}
    \caption{Comparing different models on the harder code generation tasks, CodeContest (CC)~\cite{DBLP:journals/corr/abs-2203-07814} and APPS~\cite{APPS}. DS-Instruct = DeepSeekCoder-Instruct. DS-AMR-Evol is our model.}
    \label{tab:compare_cc}
\end{table}

Since HumanEval and MBPP cover basic coding tasks, we've gone further to evaluate different models on advanced coding challenges, specifically CodeContest~\cite{DBLP:journals/corr/abs-2203-07814} and APPS~\cite{APPS}. All models generate the answers with greedy decoding. As seen in Table~\ref{tab:compare_cc}, our model not only performs better overall but also beats the official instruction model, despite it being trained on much more data than ours.
\section{Conclusion}

In this study, we present a novel framework, \textbf{AMR-Evol}, that leverages a two-stage approach—namely, modular decomposition and adaptive response evolution—to enhance code response distillation from teacher models, thereby improving knowledge distillation in code generation. Our experiments across three well-known coding benchmarks, HumanEval, MBPP, and EvalPlus, demonstrate the effectiveness of our method.

\section*{Acknowledgement}
This work is partially supported by National Natural Science Foundation of China Young Scientists Fund(No. 62206233) and Hong Kong RGC ECS (No. 22200722).
\section*{Limitation}

Our framework has room for enhancement in several aspects:
\begin{itemize}
\item First, despite Figure~\ref{fig:human} showcasing our method's capacity to improve the accuracy of code response distillation, achieving 100\% accuracy remains unattainable. While our approach does alleviate this concern to some extent, the risk of delivering low-quality responses that could potentially mislead the student models cannot be entirely eliminated. Future endeavors could explore the integration of tools, such as compilers, to further refine the quality of the responses.

\item Second, our framework's enhanced capability for code knowledge distillation is accompanied by a requirement for multi-stage generation, leading to increased costs in leveraging the teacher models. This cost-performance trade-off has been discussed in Appendix~\ref{app:cost}, where we conclude that the benefits in performance outweigh the incremental costs incurred.

\item Third, the design of our method is narrowly focused on code knowledge distillation, limiting its broader application across general domains. The foundation of our framework in modular programming principles presents considerable obstacles in adapting its method for use in non-coding areas.
\end{itemize}

\bibliography{custom}

\begin{thebibliography}{62}
\providecommand{\natexlab}[1]{#1}

\bibitem[{Anil et~al.(2023{\natexlab{a}})Anil, Borgeaud, Wu, Alayrac, Yu, Soricut, Schalkwyk, Dai, Hauth, Millican, Silver, Petrov, Johnson, Antonoglou, Schrittwieser, Glaese, Chen, Pitler, Lillicrap, Lazaridou, Firat, Molloy, Isard, Barham, Hennigan, Lee, Viola, Reynolds, Xu, Doherty, Collins, Meyer, Rutherford, Moreira, Ayoub, Goel, Tucker, Piqueras, Krikun, Barr, Savinov, Danihelka, Roelofs, White, Andreassen, von Glehn, Yagati, Kazemi, Gonzalez, Khalman, Sygnowski, and et~al.}]{gemini}
Rohan Anil, Sebastian Borgeaud, Yonghui Wu, Jean{-}Baptiste Alayrac, Jiahui Yu, Radu Soricut, Johan Schalkwyk, Andrew~M. Dai, Anja Hauth, Katie Millican, David Silver, Slav Petrov, Melvin Johnson, Ioannis Antonoglou, Julian Schrittwieser, Amelia Glaese, Jilin Chen, Emily Pitler, Timothy~P. Lillicrap, Angeliki Lazaridou, Orhan Firat, James Molloy, Michael Isard, Paul~Ronald Barham, Tom Hennigan, Benjamin Lee, Fabio Viola, Malcolm Reynolds, Yuanzhong Xu, Ryan Doherty, Eli Collins, Clemens Meyer, Eliza Rutherford, Erica Moreira, Kareem Ayoub, Megha Goel, George Tucker, Enrique Piqueras, Maxim Krikun, Iain Barr, Nikolay Savinov, Ivo Danihelka, Becca Roelofs, Ana{\"{\i}}s White, Anders Andreassen, Tamara von Glehn, Lakshman Yagati, Mehran Kazemi, Lucas Gonzalez, Misha Khalman, Jakub Sygnowski, and et~al. 2023{\natexlab{a}}.
\newblock \href {https://doi.org/10.48550/ARXIV.2312.11805} {Gemini: {A} family of highly capable multimodal models}.
\newblock \emph{CoRR}, abs/2312.11805.

\bibitem[{Anil et~al.(2023{\natexlab{b}})Anil, Dai, Firat, Johnson, Lepikhin, Passos, Shakeri, Taropa, Bailey, Chen, Chu, Clark, Shafey, Huang, Meier{-}Hellstern, Mishra, Moreira, Omernick, Robinson, Ruder, Tay, Xiao, Xu, Zhang, {\'{A}}brego, Ahn, Austin, Barham, Botha, Bradbury, Brahma, Brooks, Catasta, Cheng, Cherry, Choquette{-}Choo, Chowdhery, Crepy, Dave, Dehghani, Dev, Devlin, D{\'{\i}}az, Du, Dyer, Feinberg, Feng, Fienber, Freitag, Garcia, Gehrmann, Gonzalez, and et~al.}]{palm2}
Rohan Anil, Andrew~M. Dai, Orhan Firat, Melvin Johnson, Dmitry Lepikhin, Alexandre Passos, Siamak Shakeri, Emanuel Taropa, Paige Bailey, Zhifeng Chen, Eric Chu, Jonathan~H. Clark, Laurent~El Shafey, Yanping Huang, Kathy Meier{-}Hellstern, Gaurav Mishra, Erica Moreira, Mark Omernick, Kevin Robinson, Sebastian Ruder, Yi~Tay, Kefan Xiao, Yuanzhong Xu, Yujing Zhang, Gustavo~Hern{\'{a}}ndez {\'{A}}brego, Junwhan Ahn, Jacob Austin, Paul Barham, Jan~A. Botha, James Bradbury, Siddhartha Brahma, Kevin Brooks, Michele Catasta, Yong Cheng, Colin Cherry, Christopher~A. Choquette{-}Choo, Aakanksha Chowdhery, Cl{\'{e}}ment Crepy, Shachi Dave, Mostafa Dehghani, Sunipa Dev, Jacob Devlin, Mark D{\'{\i}}az, Nan Du, Ethan Dyer, Vladimir Feinberg, Fangxiaoyu Feng, Vlad Fienber, Markus Freitag, Xavier Garcia, Sebastian Gehrmann, Lucas Gonzalez, and et~al. 2023{\natexlab{b}}.
\newblock \href {https://doi.org/10.48550/arXiv.2305.10403} {Palm 2 technical report}.
\newblock \emph{CoRR}, abs/2305.10403.

\bibitem[{Anthropic(2023)}]{claude}
Anthropic. 2023.
\newblock Claude: A family of large language models.
\newblock \url{https://www.anthropic.com/claude}.

\bibitem[{Austin et~al.(2021)Austin, Odena, Nye, Bosma, Michalewski, Dohan, Jiang, Cai, Terry, Le, and Sutton}]{MBPP}
Jacob Austin, Augustus Odena, Maxwell~I. Nye, Maarten Bosma, Henryk Michalewski, David Dohan, Ellen Jiang, Carrie~J. Cai, Michael Terry, Quoc~V. Le, and Charles Sutton. 2021.
\newblock \href {https://arxiv.org/abs/2108.07732} {Program synthesis with large language models}.
\newblock \emph{CoRR}, abs/2108.07732.

\bibitem[{Bai et~al.(2023)Bai, Bai, Chu, Cui, Dang, Deng, Fan, Ge, Han, Huang, Hui, Ji, Li, Lin, Lin, Liu, Liu, Lu, Lu, Ma, Men, Ren, Ren, Tan, Tan, Tu, Wang, Wang, Wang, Wu, Xu, Xu, Yang, Yang, Yang, Yang, Yao, Yu, Yuan, Yuan, Zhang, Zhang, Zhang, Zhang, Zhou, Zhou, Zhou, and Zhu}]{qwen}
Jinze Bai, Shuai Bai, Yunfei Chu, Zeyu Cui, Kai Dang, Xiaodong Deng, Yang Fan, Wenbin Ge, Yu~Han, Fei Huang, Binyuan Hui, Luo Ji, Mei Li, Junyang Lin, Runji Lin, Dayiheng Liu, Gao Liu, Chengqiang Lu, Keming Lu, Jianxin Ma, Rui Men, Xingzhang Ren, Xuancheng Ren, Chuanqi Tan, Sinan Tan, Jianhong Tu, Peng Wang, Shijie Wang, Wei Wang, Shengguang Wu, Benfeng Xu, Jin Xu, An~Yang, Hao Yang, Jian Yang, Shusheng Yang, Yang Yao, Bowen Yu, Hongyi Yuan, Zheng Yuan, Jianwei Zhang, Xingxuan Zhang, Yichang Zhang, Zhenru Zhang, Chang Zhou, Jingren Zhou, Xiaohuan Zhou, and Tianhang Zhu. 2023.
\newblock Qwen technical report.
\newblock \emph{arXiv preprint arXiv:2309.16609}.

\bibitem[{Brown et~al.(2020)Brown, Mann, Ryder, Subbiah, Kaplan, Dhariwal, Neelakantan, Shyam, Sastry, Askell, Agarwal, Herbert{-}Voss, Krueger, Henighan, Child, Ramesh, Ziegler, Wu, Winter, Hesse, Chen, Sigler, Litwin, Gray, Chess, Clark, Berner, McCandlish, Radford, Sutskever, and Amodei}]{gpt3}
Tom~B. Brown, Benjamin Mann, Nick Ryder, Melanie Subbiah, Jared Kaplan, Prafulla Dhariwal, Arvind Neelakantan, Pranav Shyam, Girish Sastry, Amanda Askell, Sandhini Agarwal, Ariel Herbert{-}Voss, Gretchen Krueger, Tom Henighan, Rewon Child, Aditya Ramesh, Daniel~M. Ziegler, Jeffrey Wu, Clemens Winter, Christopher Hesse, Mark Chen, Eric Sigler, Mateusz Litwin, Scott Gray, Benjamin Chess, Jack Clark, Christopher Berner, Sam McCandlish, Alec Radford, Ilya Sutskever, and Dario Amodei. 2020.
\newblock \href {https://proceedings.neurips.cc/paper/2020/hash/1457c0d6bfcb4967418bfb8ac142f64a-Abstract.html} {Language models are few-shot learners}.
\newblock In \emph{Advances in Neural Information Processing Systems 33: Annual Conference on Neural Information Processing Systems 2020, NeurIPS 2020, December 6-12, 2020, virtual}.

\bibitem[{Chaudhary(2023)}]{codealpaca}
Sahil Chaudhary. 2023.
\newblock Code alpaca: An instruction-following llama model for code generation.
\newblock \url{https://github.com/sahil280114/codealpaca}.

\bibitem[{Chen et~al.(2023{\natexlab{a}})Chen, Zhang, Nguyen, Zan, Lin, Lou, and Chen}]{CodeT}
Bei Chen, Fengji Zhang, Anh Nguyen, Daoguang Zan, Zeqi Lin, Jian{-}Guang Lou, and Weizhu Chen. 2023{\natexlab{a}}.
\newblock \href {https://openreview.net/pdf?id=ktrw68Cmu9c} {Codet: Code generation with generated tests}.
\newblock In \emph{The Eleventh International Conference on Learning Representations, {ICLR} 2023, Kigali, Rwanda, May 1-5, 2023}. OpenReview.net.

\bibitem[{Chen et~al.(2023{\natexlab{b}})Chen, Saha, Hoi, and Joty}]{DBLP:journals/corr/abs-2310-18628}
Hailin Chen, Amrita Saha, Steven C.~H. Hoi, and Shafiq Joty. 2023{\natexlab{b}}.
\newblock \href {https://doi.org/10.48550/ARXIV.2310.18628} {Personalised distillation: Empowering open-sourced llms with adaptive learning for code generation}.
\newblock \emph{CoRR}, abs/2310.18628.

\bibitem[{Chen et~al.(2021)Chen, Tworek, Jun, Yuan, de~Oliveira~Pinto, Kaplan, Edwards, Burda, Joseph, Brockman, Ray, Puri, Krueger, Petrov, Khlaaf, Sastry, Mishkin, Chan, Gray, Ryder, Pavlov, Power, Kaiser, Bavarian, Winter, Tillet, Such, Cummings, Plappert, Chantzis, Barnes, Herbert{-}Voss, Guss, Nichol, Paino, Tezak, Tang, Babuschkin, Balaji, Jain, Saunders, Hesse, Carr, Leike, Achiam, Misra, Morikawa, Radford, Knight, Brundage, Murati, Mayer, Welinder, McGrew, Amodei, McCandlish, Sutskever, and Zaremba}]{humeval}
Mark Chen, Jerry Tworek, Heewoo Jun, Qiming Yuan, Henrique~Pond{\'{e}} de~Oliveira~Pinto, Jared Kaplan, Harrison Edwards, Yuri Burda, Nicholas Joseph, Greg Brockman, Alex Ray, Raul Puri, Gretchen Krueger, Michael Petrov, Heidy Khlaaf, Girish Sastry, Pamela Mishkin, Brooke Chan, Scott Gray, Nick Ryder, Mikhail Pavlov, Alethea Power, Lukasz Kaiser, Mohammad Bavarian, Clemens Winter, Philippe Tillet, Felipe~Petroski Such, Dave Cummings, Matthias Plappert, Fotios Chantzis, Elizabeth Barnes, Ariel Herbert{-}Voss, William~Hebgen Guss, Alex Nichol, Alex Paino, Nikolas Tezak, Jie Tang, Igor Babuschkin, Suchir Balaji, Shantanu Jain, William Saunders, Christopher Hesse, Andrew~N. Carr, Jan Leike, Joshua Achiam, Vedant Misra, Evan Morikawa, Alec Radford, Matthew Knight, Miles Brundage, Mira Murati, Katie Mayer, Peter Welinder, Bob McGrew, Dario Amodei, Sam McCandlish, Ilya Sutskever, and Wojciech Zaremba. 2021.
\newblock \href {https://arxiv.org/abs/2107.03374} {Evaluating large language models trained on code}.
\newblock \emph{CoRR}, abs/2107.03374.

\bibitem[{Chen et~al.(2023{\natexlab{c}})Chen, Lin, Sch{\"{a}}rli, and Zhou}]{DBLP:journals/corr/abs-2304-05128}
Xinyun Chen, Maxwell Lin, Nathanael Sch{\"{a}}rli, and Denny Zhou. 2023{\natexlab{c}}.
\newblock \href {https://doi.org/10.48550/ARXIV.2304.05128} {Teaching large language models to self-debug}.
\newblock \emph{CoRR}, abs/2304.05128.

\bibitem[{Chen et~al.(2023{\natexlab{d}})Chen, Lin, Sch{\"{a}}rli, and Zhou}]{self-debug}
Xinyun Chen, Maxwell Lin, Nathanael Sch{\"{a}}rli, and Denny Zhou. 2023{\natexlab{d}}.
\newblock \href {https://doi.org/10.48550/ARXIV.2304.05128} {Teaching large language models to self-debug}.
\newblock \emph{CoRR}, abs/2304.05128.

\bibitem[{Chowdhery et~al.(2022)Chowdhery, Narang, Devlin, Bosma, Mishra, Roberts, Barham, Chung, Sutton, Gehrmann, Schuh, Shi, Tsvyashchenko, Maynez, Rao, Barnes, Tay, Shazeer, Prabhakaran, Reif, Du, Hutchinson, Pope, Bradbury, Austin, Isard, Gur{-}Ari, Yin, Duke, Levskaya, Ghemawat, Dev, Michalewski, Garcia, Misra, Robinson, Fedus, Zhou, Ippolito, Luan, Lim, Zoph, Spiridonov, Sepassi, Dohan, Agrawal, Omernick, Dai, Pillai, Pellat, Lewkowycz, Moreira, Child, Polozov, Lee, Zhou, Wang, Saeta, Diaz, Firat, Catasta, Wei, Meier{-}Hellstern, Eck, Dean, Petrov, and Fiedel}]{PaLM}
Aakanksha Chowdhery, Sharan Narang, Jacob Devlin, Maarten Bosma, Gaurav Mishra, Adam Roberts, Paul Barham, Hyung~Won Chung, Charles Sutton, Sebastian Gehrmann, Parker Schuh, Kensen Shi, Sasha Tsvyashchenko, Joshua Maynez, Abhishek Rao, Parker Barnes, Yi~Tay, Noam Shazeer, Vinodkumar Prabhakaran, Emily Reif, Nan Du, Ben Hutchinson, Reiner Pope, James Bradbury, Jacob Austin, Michael Isard, Guy Gur{-}Ari, Pengcheng Yin, Toju Duke, Anselm Levskaya, Sanjay Ghemawat, Sunipa Dev, Henryk Michalewski, Xavier Garcia, Vedant Misra, Kevin Robinson, Liam Fedus, Denny Zhou, Daphne Ippolito, David Luan, Hyeontaek Lim, Barret Zoph, Alexander Spiridonov, Ryan Sepassi, David Dohan, Shivani Agrawal, Mark Omernick, Andrew~M. Dai, Thanumalayan~Sankaranarayana Pillai, Marie Pellat, Aitor Lewkowycz, Erica Moreira, Rewon Child, Oleksandr Polozov, Katherine Lee, Zongwei Zhou, Xuezhi Wang, Brennan Saeta, Mark Diaz, Orhan Firat, Michele Catasta, Jason Wei, Kathy Meier{-}Hellstern, Douglas Eck, Jeff Dean, Slav Petrov, and Noah Fiedel.
  2022.
\newblock \href {https://doi.org/10.48550/arXiv.2204.02311} {Palm: Scaling language modeling with pathways}.
\newblock \emph{CoRR}, abs/2204.02311.

\bibitem[{Dijkstra(1967)}]{DBLP:conf/sosp/Dijkstra67}
Edsger~W. Dijkstra. 1967.
\newblock \href {https://doi.org/10.1145/800001.811672} {The structure of the "the"-multiprogramming system}.
\newblock In \emph{Proceedings of the First Symposium on Operating Systems Principles, {SOSP} 1967, Gatlinburg, Tennesse, USA, 1967}. {ACM}.

\bibitem[{Google(2024)}]{codegemma}
Google. 2024.
\newblock Codegemma: Open code models based on gemma.
\newblock \url{https://storage.googleapis.com/deepmind-media/gemma/codegemma_report.pdf}.

\bibitem[{Groeneveld et~al.(2024)Groeneveld, Beltagy, Walsh, Bhagia, Kinney, Tafjord, Jha, Ivison, Magnusson, Wang, Arora, Atkinson, Authur, Chandu, Cohan, Dumas, Elazar, Gu, Hessel, Khot, Merrill, Morrison, Muennighoff, Naik, Nam, Peters, Pyatkin, Ravichander, Schwenk, Shah, Smith, Strubell, Subramani, Wortsman, Dasigi, Lambert, Richardson, Zettlemoyer, Dodge, Lo, Soldaini, Smith, and Hajishirzi}]{OLMO}
Dirk Groeneveld, Iz~Beltagy, Pete Walsh, Akshita Bhagia, Rodney Kinney, Oyvind Tafjord, Ananya~Harsh Jha, Hamish Ivison, Ian Magnusson, Yizhong Wang, Shane Arora, David Atkinson, Russell Authur, Khyathi~Raghavi Chandu, Arman Cohan, Jennifer Dumas, Yanai Elazar, Yuling Gu, Jack Hessel, Tushar Khot, William Merrill, Jacob Morrison, Niklas Muennighoff, Aakanksha Naik, Crystal Nam, Matthew~E. Peters, Valentina Pyatkin, Abhilasha Ravichander, Dustin Schwenk, Saurabh Shah, Will Smith, Emma Strubell, Nishant Subramani, Mitchell Wortsman, Pradeep Dasigi, Nathan Lambert, Kyle Richardson, Luke Zettlemoyer, Jesse Dodge, Kyle Lo, Luca Soldaini, Noah~A. Smith, and Hannaneh Hajishirzi. 2024.
\newblock \href {https://doi.org/10.48550/ARXIV.2402.00838} {Olmo: Accelerating the science of language models}.
\newblock \emph{CoRR}, abs/2402.00838.

\bibitem[{Gunasekar et~al.(2023)Gunasekar, Zhang, Aneja, Mendes, Giorno, Gopi, Javaheripi, Kauffmann, de~Rosa, Saarikivi, Salim, Shah, Behl, Wang, Bubeck, Eldan, Kalai, Lee, and Li}]{phi1}
Suriya Gunasekar, Yi~Zhang, Jyoti Aneja, Caio C{\'{e}}sar~Teodoro Mendes, Allie~Del Giorno, Sivakanth Gopi, Mojan Javaheripi, Piero Kauffmann, Gustavo de~Rosa, Olli Saarikivi, Adil Salim, Shital Shah, Harkirat~Singh Behl, Xin Wang, S{\'{e}}bastien Bubeck, Ronen Eldan, Adam~Tauman Kalai, Yin~Tat Lee, and Yuanzhi Li. 2023.
\newblock \href {https://doi.org/10.48550/ARXIV.2306.11644} {Textbooks are all you need}.
\newblock \emph{CoRR}, abs/2306.11644.

\bibitem[{Guo et~al.(2024)Guo, Zhu, Yang, Xie, Dong, Zhang, Chen, Bi, Wu, Li, Luo, Xiong, and Liang}]{deepseekcoder}
Daya Guo, Qihao Zhu, Dejian Yang, Zhenda Xie, Kai Dong, Wentao Zhang, Guanting Chen, Xiao Bi, Y.~Wu, Y.~K. Li, Fuli Luo, Yingfei Xiong, and Wenfeng Liang. 2024.
\newblock \href {https://doi.org/10.48550/ARXIV.2401.14196} {Deepseek-coder: When the large language model meets programming - the rise of code intelligence}.
\newblock \emph{CoRR}, abs/2401.14196.

\bibitem[{Hendrycks et~al.(2021)Hendrycks, Basart, Kadavath, Mazeika, Arora, Guo, Burns, Puranik, He, Song, and Steinhardt}]{APPS}
Dan Hendrycks, Steven Basart, Saurav Kadavath, Mantas Mazeika, Akul Arora, Ethan Guo, Collin Burns, Samir Puranik, Horace He, Dawn Song, and Jacob Steinhardt. 2021.
\newblock \href {https://datasets-benchmarks-proceedings.neurips.cc/paper/2021/hash/c24cd76e1ce41366a4bbe8a49b02a028-Abstract-round2.html} {Measuring coding challenge competence with {APPS}}.
\newblock In \emph{Proceedings of the Neural Information Processing Systems Track on Datasets and Benchmarks 1, NeurIPS Datasets and Benchmarks 2021, December 2021, virtual}.

\bibitem[{Hsieh et~al.(2023)Hsieh, Li, Yeh, Nakhost, Fujii, Ratner, Krishna, Lee, and Pfister}]{DBLP:conf/acl/HsiehLYNFRKLP23}
Cheng{-}Yu Hsieh, Chun{-}Liang Li, Chih{-}Kuan Yeh, Hootan Nakhost, Yasuhisa Fujii, Alex Ratner, Ranjay Krishna, Chen{-}Yu Lee, and Tomas Pfister. 2023.
\newblock \href {https://doi.org/10.18653/V1/2023.FINDINGS-ACL.507} {Distilling step-by-step! outperforming larger language models with less training data and smaller model sizes}.
\newblock In \emph{Findings of the Association for Computational Linguistics: {ACL} 2023, Toronto, Canada, July 9-14, 2023}, pages 8003--8017. Association for Computational Linguistics.

\bibitem[{Hu et~al.(2024)Hu, Tu, Han, He, Cui, Long, Zheng, Fang, Huang, Zhao, Zhang, Thai, Zhang, Wang, Yao, Zhao, Zhou, Cai, Zhai, Ding, Jia, Zeng, Li, Liu, and Sun}]{MiniCPM}
Shengding Hu, Yuge Tu, Xu~Han, Chaoqun He, Ganqu Cui, Xiang Long, Zhi Zheng, Yewei Fang, Yuxiang Huang, Weilin Zhao, Xinrong Zhang, Zhen~Leng Thai, Kai Zhang, Chongyi Wang, Yuan Yao, Chenyang Zhao, Jie Zhou, Jie Cai, Zhongwu Zhai, Ning Ding, Chao Jia, Guoyang Zeng, Dahai Li, Zhiyuan Liu, and Maosong Sun. 2024.
\newblock \href {https://doi.org/10.48550/ARXIV.2404.06395} {Minicpm: Unveiling the potential of small language models with scalable training strategies}.
\newblock \emph{CoRR}, abs/2404.06395.

\bibitem[{Jiang et~al.(2023)Jiang, Sablayrolles, Mensch, Bamford, Chaplot, de~Las~Casas, Bressand, Lengyel, Lample, Saulnier, Lavaud, Lachaux, Stock, Scao, Lavril, Wang, Lacroix, and Sayed}]{mistral}
Albert~Q. Jiang, Alexandre Sablayrolles, Arthur Mensch, Chris Bamford, Devendra~Singh Chaplot, Diego de~Las~Casas, Florian Bressand, Gianna Lengyel, Guillaume Lample, Lucile Saulnier, L{\'{e}}lio~Renard Lavaud, Marie{-}Anne Lachaux, Pierre Stock, Teven~Le Scao, Thibaut Lavril, Thomas Wang, Timoth{\'{e}}e Lacroix, and William~El Sayed. 2023.
\newblock \href {https://doi.org/10.48550/ARXIV.2310.06825} {Mistral 7b}.
\newblock \emph{CoRR}, abs/2310.06825.

\bibitem[{Le et~al.(2023)Le, Chen, Saha, Gokul, Sahoo, and Joty}]{CodeChain}
Hung Le, Hailin Chen, Amrita Saha, Akash Gokul, Doyen Sahoo, and Shafiq Joty. 2023.
\newblock \href {https://doi.org/10.48550/ARXIV.2310.08992} {Codechain: Towards modular code generation through chain of self-revisions with representative sub-modules}.
\newblock \emph{CoRR}, abs/2310.08992.

\bibitem[{Li et~al.(2024)Li, Hu, Zhao, Chen, Xie, Liu, Xie, and He}]{li2024instructcoder}
Kaixin Li, Qisheng Hu, Xu~Zhao, Hui Chen, Yuxi Xie, Tiedong Liu, Qizhe Xie, and Junxian He. 2024.
\newblock \href {https://arxiv.org/abs/2310.20329} {Instructcoder: Instruction tuning large language models for code editing}.
\newblock \emph{Preprint}, arXiv:2310.20329.

\bibitem[{Li et~al.(2023{\natexlab{a}})Li, Allal, Zi, Muennighoff, Kocetkov, Mou, Marone, Akiki, Li, Chim et~al.}]{li2023starcoder}
Raymond Li, Loubna~Ben Allal, Yangtian Zi, Niklas Muennighoff, Denis Kocetkov, Chenghao Mou, Marc Marone, Christopher Akiki, Jia Li, Jenny Chim, et~al. 2023{\natexlab{a}}.
\newblock Starcoder: may the source be with you!
\newblock \emph{arXiv preprint arXiv:2305.06161}.

\bibitem[{Li et~al.(2023{\natexlab{b}})Li, Bubeck, Eldan, Giorno, Gunasekar, and Lee}]{DBLP:journals/corr/abs-2309-05463}
Yuanzhi Li, S{\'{e}}bastien Bubeck, Ronen Eldan, Allie~Del Giorno, Suriya Gunasekar, and Yin~Tat Lee. 2023{\natexlab{b}}.
\newblock \href {https://doi.org/10.48550/ARXIV.2309.05463} {Textbooks are all you need {II:} phi-1.5 technical report}.
\newblock \emph{CoRR}, abs/2309.05463.

\bibitem[{Li et~al.(2022)Li, Choi, Chung, Kushman, Schrittwieser, Leblond, Eccles, Keeling, Gimeno, Lago, Hubert, Choy, de~Masson~d'Autume, Babuschkin, Chen, Huang, Welbl, Gowal, Cherepanov, Molloy, Mankowitz, Robson, Kohli, de~Freitas, Kavukcuoglu, and Vinyals}]{DBLP:journals/corr/abs-2203-07814}
Yujia Li, David~H. Choi, Junyoung Chung, Nate Kushman, Julian Schrittwieser, R{\'{e}}mi Leblond, Tom Eccles, James Keeling, Felix Gimeno, Agustin~Dal Lago, Thomas Hubert, Peter Choy, Cyprien de~Masson~d'Autume, Igor Babuschkin, Xinyun Chen, Po{-}Sen Huang, Johannes Welbl, Sven Gowal, Alexey Cherepanov, James Molloy, Daniel~J. Mankowitz, Esme~Sutherland Robson, Pushmeet Kohli, Nando de~Freitas, Koray Kavukcuoglu, and Oriol Vinyals. 2022.
\newblock \href {https://doi.org/10.48550/ARXIV.2203.07814} {Competition-level code generation with alphacode}.
\newblock \emph{CoRR}, abs/2203.07814.

\bibitem[{Li et~al.(2023{\natexlab{c}})Li, Zhang, Zhang, Long, Xie, and Zhang}]{li2023towards}
Zehan Li, Xin Zhang, Yanzhao Zhang, Dingkun Long, Pengjun Xie, and Meishan Zhang. 2023{\natexlab{c}}.
\newblock Towards general text embeddings with multi-stage contrastive learning.
\newblock \emph{arXiv preprint arXiv:2308.03281}.

\bibitem[{Liu et~al.(2023)Liu, Xia, Wang, and Zhang}]{humanevalp}
Jiawei Liu, Chunqiu~Steven Xia, Yuyao Wang, and Lingming Zhang. 2023.
\newblock \href {https://doi.org/10.48550/arXiv.2305.01210} {Is your code generated by chatgpt really correct? rigorous evaluation of large language models for code generation}.
\newblock \emph{CoRR}, abs/2305.01210.

\bibitem[{Liu et~al.(2024)Liu, Wei, Liu, Si, Zhang, Rao, Zheng, Peng, Yang, Zhou, and Dai}]{data_synthetic}
Ruibo Liu, Jerry Wei, Fangyu Liu, Chenglei Si, Yanzhe Zhang, Jinmeng Rao, Steven Zheng, Daiyi Peng, Diyi Yang, Denny Zhou, and Andrew~M. Dai. 2024.
\newblock \href {https://doi.org/10.48550/ARXIV.2404.07503} {Best practices and lessons learned on synthetic data for language models}.
\newblock \emph{CoRR}, abs/2404.07503.

\bibitem[{Loshchilov and Hutter(2019)}]{adamw}
Ilya Loshchilov and Frank Hutter. 2019.
\newblock \href {https://openreview.net/forum?id=Bkg6RiCqY7} {Decoupled weight decay regularization}.
\newblock In \emph{7th International Conference on Learning Representations, {ICLR} 2019, New Orleans, LA, USA, May 6-9, 2019}. OpenReview.net.

\bibitem[{Lozhkov et~al.(2024)Lozhkov, Li, Allal, Cassano, Lamy{-}Poirier, Tazi, Tang, Pykhtar, Liu, Wei, Liu, Tian, Kocetkov, Zucker, Belkada, Wang, Liu, Abulkhanov, Paul, Li, Li, Risdal, Li, Zhu, Zhuo, Zheltonozhskii, Dade, Yu, Krau{\ss}, Jain, Su, He, Dey, Abati, Chai, Muennighoff, Tang, Oblokulov, Akiki, Marone, Mou, Mishra, Gu, Hui, Dao, Zebaze, Dehaene, Patry, Xu, McAuley, Hu, Scholak, Paquet, Robinson, Anderson, Chapados, and et~al.}]{starcoder2}
Anton Lozhkov, Raymond Li, Loubna~Ben Allal, Federico Cassano, Joel Lamy{-}Poirier, Nouamane Tazi, Ao~Tang, Dmytro Pykhtar, Jiawei Liu, Yuxiang Wei, Tianyang Liu, Max Tian, Denis Kocetkov, Arthur Zucker, Younes Belkada, Zijian Wang, Qian Liu, Dmitry Abulkhanov, Indraneil Paul, Zhuang Li, Wen{-}Ding Li, Megan Risdal, Jia Li, Jian Zhu, Terry~Yue Zhuo, Evgenii Zheltonozhskii, Nii Osae~Osae Dade, Wenhao Yu, Lucas Krau{\ss}, Naman Jain, Yixuan Su, Xuanli He, Manan Dey, Edoardo Abati, Yekun Chai, Niklas Muennighoff, Xiangru Tang, Muhtasham Oblokulov, Christopher Akiki, Marc Marone, Chenghao Mou, Mayank Mishra, Alex Gu, Binyuan Hui, Tri Dao, Armel Zebaze, Olivier Dehaene, Nicolas Patry, Canwen Xu, Julian~J. McAuley, Han Hu, Torsten Scholak, S{\'{e}}bastien Paquet, Jennifer Robinson, Carolyn~Jane Anderson, Nicolas Chapados, and et~al. 2024.
\newblock \href {https://doi.org/10.48550/ARXIV.2402.19173} {Starcoder 2 and the stack v2: The next generation}.
\newblock \emph{CoRR}, abs/2402.19173.

\bibitem[{Luo et~al.(2024)Luo, Xu, Zhao, Sun, Geng, Hu, Tao, Ma, Lin, and Jiang}]{wizardcoder}
Ziyang Luo, Can Xu, Pu~Zhao, Qingfeng Sun, Xiubo Geng, Wenxiang Hu, Chongyang Tao, Jing Ma, Qingwei Lin, and Daxin Jiang. 2024.
\newblock \href {https://openreview.net/forum?id=UnUwSIgK5W} {Wizardcoder: Empowering code large language models with evol-instruct}.
\newblock In \emph{The Twelfth International Conference on Learning Representations}.

\bibitem[{Mesnard et~al.(2024)Mesnard, Hardin, Dadashi, Bhupatiraju, Pathak, Sifre, Rivi{\`{e}}re, Kale, Love, Tafti, Hussenot, Chowdhery, Roberts, Barua, Botev, Castro{-}Ros, Slone, H{\'{e}}liou, Tacchetti, Bulanova, Paterson, Tsai, Shahriari, Lan, Choquette{-}Choo, Crepy, Cer, Ippolito, Reid, Buchatskaya, Ni, Noland, Yan, Tucker, Muraru, Rozhdestvenskiy, Michalewski, Tenney, Grishchenko, Austin, Keeling, Labanowski, Lespiau, Stanway, Brennan, Chen, Ferret, Chiu, and et~al.}]{gemma}
Thomas Mesnard, Cassidy Hardin, Robert Dadashi, Surya Bhupatiraju, Shreya Pathak, Laurent Sifre, Morgane Rivi{\`{e}}re, Mihir~Sanjay Kale, Juliette Love, Pouya Tafti, L{\'{e}}onard Hussenot, Aakanksha Chowdhery, Adam Roberts, Aditya Barua, Alex Botev, Alex Castro{-}Ros, Ambrose Slone, Am{\'{e}}lie H{\'{e}}liou, Andrea Tacchetti, Anna Bulanova, Antonia Paterson, Beth Tsai, Bobak Shahriari, Charline~Le Lan, Christopher~A. Choquette{-}Choo, Cl{\'{e}}ment Crepy, Daniel Cer, Daphne Ippolito, David Reid, Elena Buchatskaya, Eric Ni, Eric Noland, Geng Yan, George Tucker, George{-}Cristian Muraru, Grigory Rozhdestvenskiy, Henryk Michalewski, Ian Tenney, Ivan Grishchenko, Jacob Austin, James Keeling, Jane Labanowski, Jean{-}Baptiste Lespiau, Jeff Stanway, Jenny Brennan, Jeremy Chen, Johan Ferret, Justin Chiu, and et~al. 2024.
\newblock \href {https://doi.org/10.48550/ARXIV.2403.08295} {Gemma: Open models based on gemini research and technology}.
\newblock \emph{CoRR}, abs/2403.08295.

\bibitem[{Meta(2024)}]{llama3}
Meta. 2024.
\newblock Introducing meta llama 3: The most capable openly available llm to date.
\newblock \url{https://ai.meta.com/blog/meta-llama-3/}.

\bibitem[{Mukherjee et~al.(2023)Mukherjee, Mitra, Jawahar, Agarwal, Palangi, and Awadallah}]{ORCA}
Subhabrata Mukherjee, Arindam Mitra, Ganesh Jawahar, Sahaj Agarwal, Hamid Palangi, and Ahmed Awadallah. 2023.
\newblock \href {https://doi.org/10.48550/ARXIV.2306.02707} {Orca: Progressive learning from complex explanation traces of {GPT-4}}.
\newblock \emph{CoRR}, abs/2306.02707.

\bibitem[{Nijkamp et~al.(2023{\natexlab{a}})Nijkamp, Hayashi, Xiong, Savarese, and Zhou}]{codegen2}
Erik Nijkamp, Hiroaki Hayashi, Caiming Xiong, Silvio Savarese, and Yingbo Zhou. 2023{\natexlab{a}}.
\newblock \href {https://doi.org/10.48550/arXiv.2305.02309} {Codegen2: Lessons for training llms on programming and natural languages}.
\newblock \emph{CoRR}, abs/2305.02309.

\bibitem[{Nijkamp et~al.(2023{\natexlab{b}})Nijkamp, Pang, Hayashi, Tu, Wang, Zhou, Savarese, and Xiong}]{codegen}
Erik Nijkamp, Bo~Pang, Hiroaki Hayashi, Lifu Tu, Huan Wang, Yingbo Zhou, Silvio Savarese, and Caiming Xiong. 2023{\natexlab{b}}.
\newblock \href {https://openreview.net/forum?id=iaYcJKpY2B_} {Codegen: An open large language model for code with multi-turn program synthesis}.
\newblock In \emph{The Eleventh International Conference on Learning Representations}.

\bibitem[{Olausson et~al.(2023)Olausson, Inala, Wang, Gao, and Solar{-}Lezama}]{self-repair}
Theo~X. Olausson, Jeevana~Priya Inala, Chenglong Wang, Jianfeng Gao, and Armando Solar{-}Lezama. 2023.
\newblock \href {https://doi.org/10.48550/ARXIV.2306.09896} {Demystifying {GPT} self-repair for code generation}.
\newblock \emph{CoRR}, abs/2306.09896.

\bibitem[{OpenAI(2023)}]{GPT4}
OpenAI. 2023.
\newblock \href {https://doi.org/10.48550/arXiv.2303.08774} {{GPT-4} technical report}.
\newblock \emph{CoRR}, abs/2303.08774.

\bibitem[{Parnas(1972)}]{DBLP:journals/cacm/Parnas72a}
David~Lorge Parnas. 1972.
\newblock \href {https://doi.org/10.1145/361598.361623} {On the criteria to be used in decomposing systems into modules}.
\newblock \emph{Commun. {ACM}}, 15(12):1053--1058.

\bibitem[{Rasley et~al.(2020)Rasley, Rajbhandari, Ruwase, and He}]{DBLP:conf/kdd/RasleyRRH20}
Jeff Rasley, Samyam Rajbhandari, Olatunji Ruwase, and Yuxiong He. 2020.
\newblock \href {https://doi.org/10.1145/3394486.3406703} {Deepspeed: System optimizations enable training deep learning models with over 100 billion parameters}.
\newblock In \emph{{KDD} '20: The 26th {ACM} {SIGKDD} Conference on Knowledge Discovery and Data Mining, Virtual Event, CA, USA, August 23-27, 2020}, pages 3505--3506. {ACM}.

\bibitem[{Rozi{\`{e}}re et~al.(2023)Rozi{\`{e}}re, Gehring, Gloeckle, Sootla, Gat, Tan, Adi, Liu, Remez, Rapin, Kozhevnikov, Evtimov, Bitton, Bhatt, Canton{-}Ferrer, Grattafiori, Xiong, D{\'{e}}fossez, Copet, Azhar, Touvron, Martin, Usunier, Scialom, and Synnaeve}]{codellama}
Baptiste Rozi{\`{e}}re, Jonas Gehring, Fabian Gloeckle, Sten Sootla, Itai Gat, Xiaoqing~Ellen Tan, Yossi Adi, Jingyu Liu, Tal Remez, J{\'{e}}r{\'{e}}my Rapin, Artyom Kozhevnikov, Ivan Evtimov, Joanna Bitton, Manish Bhatt, Cristian Canton{-}Ferrer, Aaron Grattafiori, Wenhan Xiong, Alexandre D{\'{e}}fossez, Jade Copet, Faisal Azhar, Hugo Touvron, Louis Martin, Nicolas Usunier, Thomas Scialom, and Gabriel Synnaeve. 2023.
\newblock \href {https://doi.org/10.48550/arXiv.2308.12950} {Code llama: Open foundation models for code}.
\newblock \emph{CoRR}, abs/2308.12950.

\bibitem[{Tay et~al.(2022)Tay, Dehghani, Tran, Garcia, Bahri, Schuster, Zheng, Houlsby, and Metzler}]{UL2}
Yi~Tay, Mostafa Dehghani, Vinh~Q. Tran, Xavier Garcia, Dara Bahri, Tal Schuster, Huaixiu~Steven Zheng, Neil Houlsby, and Donald Metzler. 2022.
\newblock \href {https://doi.org/10.48550/arXiv.2205.05131} {Unifying language learning paradigms}.
\newblock \emph{CoRR}, abs/2205.05131.

\bibitem[{Touvron et~al.(2023{\natexlab{a}})Touvron, Lavril, Izacard, Martinet, Lachaux, Lacroix, Rozi{\`{e}}re, Goyal, Hambro, Azhar, Rodriguez, Joulin, Grave, and Lample}]{llama}
Hugo Touvron, Thibaut Lavril, Gautier Izacard, Xavier Martinet, Marie{-}Anne Lachaux, Timoth{\'{e}}e Lacroix, Baptiste Rozi{\`{e}}re, Naman Goyal, Eric Hambro, Faisal Azhar, Aur{\'{e}}lien Rodriguez, Armand Joulin, Edouard Grave, and Guillaume Lample. 2023{\natexlab{a}}.
\newblock \href {https://doi.org/10.48550/arXiv.2302.13971} {Llama: Open and efficient foundation language models}.
\newblock \emph{CoRR}, abs/2302.13971.

\bibitem[{Touvron et~al.(2023{\natexlab{b}})Touvron, Martin, Stone, Albert, Almahairi, Babaei, Bashlykov, Batra, Bhargava, Bhosale, Bikel, Blecher, Canton{-}Ferrer, Chen, Cucurull, Esiobu, Fernandes, Fu, Fu, Fuller, Gao, Goswami, Goyal, Hartshorn, Hosseini, Hou, Inan, Kardas, Kerkez, Khabsa, Kloumann, Korenev, Koura, Lachaux, Lavril, Lee, Liskovich, Lu, Mao, Martinet, Mihaylov, Mishra, Molybog, Nie, Poulton, Reizenstein, Rungta, Saladi, Schelten, Silva, Smith, Subramanian, Tan, Tang, Taylor, Williams, Kuan, Xu, Yan, Zarov, Zhang, Fan, Kambadur, Narang, Rodriguez, Stojnic, Edunov, and Scialom}]{llama2}
Hugo Touvron, Louis Martin, Kevin Stone, Peter Albert, Amjad Almahairi, Yasmine Babaei, Nikolay Bashlykov, Soumya Batra, Prajjwal Bhargava, Shruti Bhosale, Dan Bikel, Lukas Blecher, Cristian Canton{-}Ferrer, Moya Chen, Guillem Cucurull, David Esiobu, Jude Fernandes, Jeremy Fu, Wenyin Fu, Brian Fuller, Cynthia Gao, Vedanuj Goswami, Naman Goyal, Anthony Hartshorn, Saghar Hosseini, Rui Hou, Hakan Inan, Marcin Kardas, Viktor Kerkez, Madian Khabsa, Isabel Kloumann, Artem Korenev, Punit~Singh Koura, Marie{-}Anne Lachaux, Thibaut Lavril, Jenya Lee, Diana Liskovich, Yinghai Lu, Yuning Mao, Xavier Martinet, Todor Mihaylov, Pushkar Mishra, Igor Molybog, Yixin Nie, Andrew Poulton, Jeremy Reizenstein, Rashi Rungta, Kalyan Saladi, Alan Schelten, Ruan Silva, Eric~Michael Smith, Ranjan Subramanian, Xiaoqing~Ellen Tan, Binh Tang, Ross Taylor, Adina Williams, Jian~Xiang Kuan, Puxin Xu, Zheng Yan, Iliyan Zarov, Yuchen Zhang, Angela Fan, Melanie Kambadur, Sharan Narang, Aur{\'{e}}lien Rodriguez, Robert Stojnic, Sergey Edunov,
  and Thomas Scialom. 2023{\natexlab{b}}.
\newblock \href {https://doi.org/10.48550/arXiv.2307.09288} {Llama 2: Open foundation and fine-tuned chat models}.
\newblock \emph{CoRR}, abs/2307.09288.

\bibitem[{Wang et~al.(2024)Wang, Zhang, Du, Zhang, and Chu}]{DBLP:journals/corr/abs-2402-05123}
Jiahao Wang, Bolin Zhang, Qianlong Du, Jiajun Zhang, and Dianhui Chu. 2024.
\newblock \href {https://doi.org/10.48550/ARXIV.2402.05123} {A survey on data selection for {LLM} instruction tuning}.
\newblock \emph{CoRR}, abs/2402.05123.

\bibitem[{Wang et~al.(2022)Wang, Kordi, Mishra, Liu, Smith, Khashabi, and Hajishirzi}]{wang2022self}
Yizhong Wang, Yeganeh Kordi, Swaroop Mishra, Alisa Liu, Noah~A Smith, Daniel Khashabi, and Hannaneh Hajishirzi. 2022.
\newblock Self-instruct: Aligning language model with self generated instructions.
\newblock \emph{arXiv preprint arXiv:2212.10560}.

\bibitem[{Wang et~al.(2023)Wang, Le, Gotmare, Bui, Li, and Hoi}]{CodeT5+}
Yue Wang, Hung Le, Akhilesh~Deepak Gotmare, Nghi D.~Q. Bui, Junnan Li, and Steven C.~H. Hoi. 2023.
\newblock \href {https://doi.org/10.48550/arXiv.2305.07922} {Codet5+: Open code large language models for code understanding and generation}.
\newblock \emph{CoRR}, abs/2305.07922.

\bibitem[{Wang et~al.(2021)Wang, Wang, Joty, and Hoi}]{codet5}
Yue Wang, Weishi Wang, Shafiq~R. Joty, and Steven C.~H. Hoi. 2021.
\newblock \href {https://doi.org/10.18653/v1/2021.emnlp-main.685} {Codet5: Identifier-aware unified pre-trained encoder-decoder models for code understanding and generation}.
\newblock In \emph{Proceedings of the 2021 Conference on Empirical Methods in Natural Language Processing, {EMNLP} 2021, Virtual Event / Punta Cana, Dominican Republic, 7-11 November, 2021}, pages 8696--8708. Association for Computational Linguistics.

\bibitem[{Wei et~al.(2022)Wei, Wang, Schuurmans, Bosma, Ichter, Xia, Chi, Le, and Zhou}]{cot}
Jason Wei, Xuezhi Wang, Dale Schuurmans, Maarten Bosma, Brian Ichter, Fei Xia, Ed~H. Chi, Quoc~V. Le, and Denny Zhou. 2022.
\newblock \href {http://papers.nips.cc/paper\_files/paper/2022/hash/9d5609613524ecf4f15af0f7b31abca4-Abstract-Conference.html} {Chain-of-thought prompting elicits reasoning in large language models}.
\newblock In \emph{Advances in Neural Information Processing Systems 35: Annual Conference on Neural Information Processing Systems 2022, NeurIPS 2022, New Orleans, LA, USA, November 28 - December 9, 2022}.

\bibitem[{Wei et~al.(2023)Wei, Wang, Liu, Ding, and Zhang}]{magiccoder}
Yuxiang Wei, Zhe Wang, Jiawei Liu, Yifeng Ding, and Lingming Zhang. 2023.
\newblock \href {https://doi.org/10.48550/ARXIV.2312.02120} {Magicoder: Source code is all you need}.
\newblock \emph{CoRR}, abs/2312.02120.

\bibitem[{Wu et~al.(2023)Wu, Duan, and Ni}]{gpt_privacy}
Xiaodong Wu, Ran Duan, and Jianbing Ni. 2023.
\newblock \href {https://doi.org/10.48550/ARXIV.2307.14192} {Unveiling security, privacy, and ethical concerns of chatgpt}.
\newblock \emph{CoRR}, abs/2307.14192.

\bibitem[{Xia et~al.(2024)Xia, Wang, Liu, Li, Yu, Chen, McAuley, and Li}]{Chain-of-X}
Yu~Xia, Rui Wang, Xu~Liu, Mingyan Li, Tong Yu, Xiang Chen, Julian McAuley, and Shuai Li. 2024.
\newblock \href {https://api.semanticscholar.org/CorpusID:269330085} {Beyond chain-of-thought: A survey of chain-of-x paradigms for llms}.

\bibitem[{Xu et~al.(2024)Xu, Li, Tao, Shen, Cheng, Li, Xu, Tao, and Zhou}]{KD_LLM}
Xiaohan Xu, Ming Li, Chongyang Tao, Tao Shen, Reynold Cheng, Jinyang Li, Can Xu, Dacheng Tao, and Tianyi Zhou. 2024.
\newblock \href {https://doi.org/10.48550/ARXIV.2402.13116} {A survey on knowledge distillation of large language models}.
\newblock \emph{CoRR}, abs/2402.13116.

\bibitem[{Yu et~al.(2023)Yu, Zhang, Shang, Huang, Xu, Zhao, Hu, and Yin}]{wavecoder}
Zhaojian Yu, Xin Zhang, Ning Shang, Yangyu Huang, Can Xu, Yishujie Zhao, Wenxiang Hu, and Qiufeng Yin. 2023.
\newblock \href {https://doi.org/10.48550/ARXIV.2312.14187} {Wavecoder: Widespread and versatile enhanced instruction tuning with refined data generation}.
\newblock \emph{CoRR}, abs/2312.14187.

\bibitem[{Yuan et~al.(2024)Yuan, Cui, Wang, Ding, Wang, Deng, Shan, Chen, Xie, Lin, Liu, Zhou, Peng, Liu, and Sun}]{yuan2024advancing}
Lifan Yuan, Ganqu Cui, Hanbin Wang, Ning Ding, Xingyao Wang, Jia Deng, Boji Shan, Huimin Chen, Ruobing Xie, Yankai Lin, Zhenghao Liu, Bowen Zhou, Hao Peng, Zhiyuan Liu, and Maosong Sun. 2024.
\newblock \href {https://arxiv.org/abs/2404.02078} {Advancing llm reasoning generalists with preference trees}.
\newblock \emph{Preprint}, arXiv:2404.02078.

\bibitem[{Zeng et~al.(2022)Zeng, Liu, Du, Wang, Lai, Ding, Yang, Xu, Zheng, Xia, Tam, Ma, Xue, Zhai, Chen, Zhang, Dong, and Tang}]{GLM-130B}
Aohan Zeng, Xiao Liu, Zhengxiao Du, Zihan Wang, Hanyu Lai, Ming Ding, Zhuoyi Yang, Yifan Xu, Wendi Zheng, Xiao Xia, Weng~Lam Tam, Zixuan Ma, Yufei Xue, Jidong Zhai, Wenguang Chen, Peng Zhang, Yuxiao Dong, and Jie Tang. 2022.
\newblock \href {https://doi.org/10.48550/arXiv.2210.02414} {{GLM-130B:} an open bilingual pre-trained model}.
\newblock \emph{CoRR}, abs/2210.02414.

\bibitem[{Zhang et~al.(2022)Zhang, Roller, Goyal, Artetxe, Chen, Chen, Dewan, Diab, Li, Lin, Mihaylov, Ott, Shleifer, Shuster, Simig, Koura, Sridhar, Wang, and Zettlemoyer}]{opt}
Susan Zhang, Stephen Roller, Naman Goyal, Mikel Artetxe, Moya Chen, Shuohui Chen, Christopher Dewan, Mona~T. Diab, Xian Li, Xi~Victoria Lin, Todor Mihaylov, Myle Ott, Sam Shleifer, Kurt Shuster, Daniel Simig, Punit~Singh Koura, Anjali Sridhar, Tianlu Wang, and Luke Zettlemoyer. 2022.
\newblock \href {https://doi.org/10.48550/arXiv.2205.01068} {{OPT:} open pre-trained transformer language models}.
\newblock \emph{CoRR}, abs/2205.01068.

\bibitem[{Zheng et~al.(2023)Zheng, Xia, Zou, Dong, Wang, Xue, Wang, Shen, Wang, Li, Su, Yang, and Tang}]{CodeGeeX}
Qinkai Zheng, Xiao Xia, Xu~Zou, Yuxiao Dong, Shan Wang, Yufei Xue, Zihan Wang, Lei Shen, Andi Wang, Yang Li, Teng Su, Zhilin Yang, and Jie Tang. 2023.
\newblock \href {https://doi.org/10.48550/arXiv.2303.17568} {Codegeex: {A} pre-trained model for code generation with multilingual evaluations on humaneval-x}.
\newblock \emph{CoRR}, abs/2303.17568.

\bibitem[{Zheng et~al.(2024)Zheng, Zhang, Shen, Liu, Lin, Fu, Chen, and Yue}]{DBLP:journals/corr/abs-2402-14658}
Tianyu Zheng, Ge~Zhang, Tianhao Shen, Xueling Liu, Bill~Yuchen Lin, Jie Fu, Wenhu Chen, and Xiang Yue. 2024.
\newblock \href {https://doi.org/10.48550/ARXIV.2402.14658} {Opencodeinterpreter: Integrating code generation with execution and refinement}.
\newblock \emph{CoRR}, abs/2402.14658.

\bibitem[{Zhou et~al.(2023)Zhou, Liu, Xu, Iyer, Sun, Mao, Ma, Efrat, Yu, Yu, Zhang, Ghosh, Lewis, Zettlemoyer, and Levy}]{LIMA}
Chunting Zhou, Pengfei Liu, Puxin Xu, Srinivasan Iyer, Jiao Sun, Yuning Mao, Xuezhe Ma, Avia Efrat, Ping Yu, Lili Yu, Susan Zhang, Gargi Ghosh, Mike Lewis, Luke Zettlemoyer, and Omer Levy. 2023.
\newblock \href {http://papers.nips.cc/paper\_files/paper/2023/hash/ac662d74829e4407ce1d126477f4a03a-Abstract-Conference.html} {{LIMA:} less is more for alignment}.
\newblock In \emph{Advances in Neural Information Processing Systems 36: Annual Conference on Neural Information Processing Systems 2023, NeurIPS 2023, New Orleans, LA, USA, December 10 - 16, 2023}.

\end{thebibliography}

\clearpage

\appendix

\section{Baselines}~\label{app:baseline}

To ensure a fair comparison, we incorporate three distinct response distillation methods as our baselines. The first method is \textbf{direct} distillation. As outlined in Section~\ref{sec:direct}, this approach involves using the teacher model to directly produce responses based on the provided code instructions. The prompt used is as follows:
\newenvironment{myblock1}{%
  \begin{tcolorbox}[colback=beaublue!8!white,colframe=beaublue!10!black,title=Prompt for Direct Distillation]
}{%
  \end{tcolorbox}
}
\begin{myblock1}
\textbf{System}: You are a professional coder. Your answer must include Python code in Markdown format.\\\\
\textbf{User}: \{instruction\}
\end{myblock1}
The second method involves response distillation utilizing the Chain-of-Thought (\textbf{CoT}) approach. We adopt the method from the few-shot CoT~\cite{cot}, prompting the teacher model to produce the responses. To minimize costs, we opt to include a single example in our prompt:
\newenvironment{myblock2}{%
  \begin{tcolorbox}[colback=beaublue!8!white,colframe=beaublue!10!black,title=Prompt for CoT Distillation]
}{%
  \end{tcolorbox}
}
\begin{myblock2}
\textbf{System}: You are a professional coder. You will be given a Python Question. Your objective is to develop an accurate solution to the Python Question. Begin by step-by-step think about your approach to solve this question, then proceed to generate your final code response in Markdown format.\\

\#\# One-Shot Example\\
\#\#\# Python Question:\\
\{one-shot-example-question\}\\

\#\#\# Correct Solution:\\
\{one-shot-example-solution\}\\

\textbf{User}: \#\# New Task\\
\#\#\# Python Question:\\
\{question\}\\

\#\#\# Correct Solution:
\end{myblock2}
\begin{table}
    \small
    \centering
    \begin{tabular}{lcc}
        \toprule
        \textbf{Data} & \textbf{Source} & \textbf{Number}\\
        \midrule
        \midrule
        \textbf{Seed} & MBPP-Train & 332\\
        \textbf{Self-Instruct} & Seed & 10k\\
        \textbf{Complex 1} & Self-Instruct & 9.8k\\
        \textbf{Complex 2} & Complex 1 & 9.7k\\
        \textbf{Complex 3} & Complex 2 & 9.7k\\
        \bottomrule
    \end{tabular}
    \caption{Statistics of Our Instruction Dataset.}
    \label{tab:ins_statistics}
\end{table}
\begin{table}
    \small
    \centering
    \begin{tabular}{lcc}
        \toprule
        \textbf{Data} & \textbf{\#Question} & \textbf{\#Avg. Tests}\\
        \midrule
        \midrule
        \textbf{HumanEval} & 164 & 9.6\\
        \textbf{HumanEval-Plus} & 164 & x80\\
        \textbf{MBPP} & 399 & 3\\
        \textbf{MBPP-Plus} & 399 & x35\\
        \bottomrule
    \end{tabular}
    \caption{Statistics of Our Benchmarks.}
    \label{tab:benchmark_statistics}
\end{table}
The third baseline, \textbf{AnsRepair}, incorporates self-repair techniques~\cite{CodeT,self-repair}. This method employs the teacher model to generate unit test functions for each sample, enabling the model to verify the correctness of its own answers. The employed prompt is as follows:
\newenvironment{myblock3}{%
  \begin{tcolorbox}[colback=beaublue!8!white,colframe=beaublue!10!black,title=Prompt for Test Function Generation]
}{%
  \end{tcolorbox}
}
\begin{myblock3}
\textbf{System}: You are a professional coder. You will be given a Python Question and its possible code solution. Your objective is to provide a test function to test whether the code solution is correct or not. Your response should be in Markdown format.\\

\#\# One-Shot Example\\
\#\#\# Python Question:\\
\{one-shot-example-question\}\\

\#\#\# Possible Code Solution:\\
\{one-shot-example-solution\}\\

\#\#\# Tests Function:\\
\{one-shot-example-tests\}\\

\textbf{User}: \#\# New Task\\
\#\#\# Python Question:\\
\{question\}\\

\#\#\# Possible Code Solution:\\
\{answer\}\\

\#\#\# Tests Function:
\end{myblock3}
Upon obtaining the test functions for each sample, we execute these tests to assess the output's correctness. Should the output fail to meet the criteria set by the test functions, we prompt the teacher model to regenerate the output. The prompt used for this process is as follows:

\newenvironment{myblock4}{%
  \begin{tcolorbox}[colback=beaublue!8!white,colframe=beaublue!10!black,title=Prompt for AnsRepair Distillation]
}{%
  \end{tcolorbox}
}
\begin{myblock4}
\textbf{System}: You are a professional coder. You will be given a Python Question and its wrong solution. You need to provide the correct solution for the Python Question in Markdown format.\\

\#\# One-Shot Example\\
\#\#\# Python Question:\\
\{one-shot-example-question\}\\

\#\#\# Wrong Solution:\\
\{one-shot-example-wrong-answer\}\\

\#\#\# Correct Solution:\\
\{one-shot-example-correct-answer\}\\

\textbf{User}: \#\# New Task\\
\#\#\# Python Question:\\
\{question\}\\

\#\#\# Wrong Solution:\\
\{answer\}\\

\#\#\# Correct Solution:
\end{myblock4}

\section{Datasets}~\label{app:datasets}

Our framework concentrates on distilling responses and requires a dataset of instructions for this purpose. As indicated in Table~\ref{tab:ins_statistics}, we enumerate the quantity of instructions used in our experiments. We initiate our process with the MBPP training set (task-ids 601-974) as a seed dataset, which enhances our ability to generate Python code effectively. To prevent any overlap with the EvalPlus test data, we are diligent in omitting any samples that coincide with the test set, thereby narrowing our training set to 332 unique MBPP tasks. We then utilize this filtered seed data and apply the self-instruction method to construct instructions. Subsequently, we employ the Code Evol-Instruct method to iteratively generate instructions of varying complexity across three distinct levels.

To ensure decontamination of our datasets, we invoke a method akin to the work of Code Evol-Instruct~\cite{wizardcoder} for data filtering. This involves employing the \texttt{gte-large-en-v1.5} model to treat each test set sample as a query, which retrieves the top five most similar samples from the training data. Subsequently, these pairs are evaluated by GPT4 in a binary classification task to decide whether a match exists. Detected matches lead to the exclusion of those specific training samples to eliminate potential data leakage.

\newenvironment{myblock5}{%
  \begin{tcolorbox}[colback=beaublue!8!white,colframe=beaublue!10!black,title=Prompt for Modular Decomposition]
}{%
  \end{tcolorbox}
}
\begin{myblock5}
\textbf{System}: You will be presented with a Python coding question along with a potential solution. Your task is to deconstruct the given solution into smaller, manageable modules. Each module should be clearly defined with specific function names, detailed input/output specifications, and concise function descriptions. Do NOT repeat the functions in the One-Shot Example.\\

\#\# One-Shot Example\\
\#\#\# Python Question:\\
\{one-shot-example-question\}\\

\#\#\# Potential Solution:\\
\{one-shot-example-solution\}\\

\#\#\# RESPONSE:\\
\{one-shot-example-modules\}\\

\textbf{User}: \#\# New Task\\
\#\#\# Python Question:\\
\{question\}\\

\#\#\# Potential Solution:\\
\{answer\}\\

\#\#\# RESPONSE:
\end{myblock5}

\section{Benchmark}~\label{app:benchmark}

Table~\ref{tab:benchmark_statistics} details the quantity of questions along with the average number of unit tests per question across all the benchmarks utilized in our study. The license of HumanEval is MIT.\footnote{\url{https://huggingface.co/datasets/openai/openai_humaneval}} The license of MBPP is cc-by-4.0.\footnote{\url{https://huggingface.co/datasets/google-research-datasets/mbpp}} The license of EvalPlus is Apache-2.0.\footnote{\url{https://github.com/evalplus/evalplus}}

\section{Implementation Details}~\label{app:impl}

Our \textbf{AMR-Evol} framework encompasses a two-stage process. In the first stage, Modular Decomposition is applied to break down the code instructions into multiple sub-modules, using the direct responses as the initial seed data. The prompt utilized for this stage is demonstrated above. During the second stage, Adaptive Response Evolution refines these decomposed sub-modules, utilizing the retrieved modules to develop the final answer. The corresponding prompt for this stage is as follows:
\newenvironment{myblock6}{%
  \begin{tcolorbox}[colback=beaublue!8!white,colframe=beaublue!10!black,title=Prompt for Adaptive Response Evolution]
}{%
  \end{tcolorbox}
}
\begin{myblock6}
\textbf{System}: You are a professional coder. You will be given a Python Question and a selection of relevant, modularized functions intended to inspire your approach. Your objective is to develop a more refined and accurate solution to the Python Question. Your response should pretend that you have never seen the Relevant Functions.\\

\#\# One-Shot Example\\
\#\#\# Python Question:\\
\{one-shot-example-question\}\\

\#\#\# Relevant Functions:\\
\{one-shot-example-similar-functions\}\\

\#\#\# Correct Solution:\\
\{one-shot-example-solution\}\\

\textbf{User}: \#\# New Task\\
\#\#\# Python Question:\\
\{question\}\\

\#\#\# Relevant Functions:\\
\{similar-functions\}\\

\#\#\# Correct Solution:
\end{myblock6}

For all instruction construction processes, we set the temperature to 0.7 and the sequence length to 2048. For all response distillation processes, the temperature is fixed at 0.0, and the sequence length is set to 3000. We train the models for 200 steps across 3 epochs with a sequence length of 2048, employing the AdamW optimizer, BF16 precision, and DeepSpeed Zero-2~\cite{DBLP:conf/kdd/RasleyRRH20}. The training is conducted on 4 A800 GPUs.

\begin{table*}
    \small
    \centering
    \begin{tabular}{lcccc}
        \toprule
        \textbf{Model} & \textbf{HE-Plus (Pass@1)} & \textbf{HE-Plus (Pass@10)} & \textbf{MBPP-Plus (Pass@1)} & \textbf{MBPP-Plus (Pass@10)}\\
        \midrule
        \midrule
        \textbf{MagiCoder-DS} & 56.0 & 72.5 & 61.7 & 68.5\\
        \textbf{WaveCoder-DS} & 56.6 & 63.2 & 57.6 & 63.0\\
        \textbf{DS-AMR-Evol} & 59.1 & 75.2 & 61.3 & 70.7\\
        \bottomrule
    \end{tabular}
    \caption{Results of pass@k(\%) on HE-Plus, MBPP-Plus. We follow the previous works~\citep{humeval} to generate n=200 samples to estimate the pass@k scores our models with the same set of hyper-parameters: temperate=0.2, and top\_p=0.95. DS-AMR-Evol is our model.}
    \label{tab:compare_sampling}
\end{table*}
\begin{table}
    \small
    \centering
    \begin{tabular}{lcc}
        \toprule
        \textbf{Teacher} & \textbf{HE-Plus} & \textbf{MBPP-Plus}\\
        \midrule
        \midrule
        GPT3.5-Turbo & 61.0 & 62.9\\
        Llama-3-70B & 62.2 & 63.2\\
        \bottomrule
    \end{tabular}
    \caption{Adopting open-source model, Llama-3-70B-Instruct, as our teacher model.}
    \label{tab:open_teacher}
\end{table}

\section{Qualitative Comparison}~\label{app:qualitative}

Table~\ref{tab:example1}~\ref{tab:example2}~\ref{tab:example3} display distilled responses obtained through various methods. It is evident from the comparison that our framework facilitates the generation of better responses for code knowledge distillation.

\begin{table*}
    \small
    \centering
    \begin{tabular}{p{0.05\textwidth}|p{0.9\textwidth}}
        \toprule
        Ins. & Design a function that implements the Laplace expansion theorem to calculate the determinant of a 3x3 matrix, but now with the added condition of utilizing nested for loops and ensuring that the matrix elements are limited to integers only.\\
        \midrule
        Direct &
        \begin{minipage}{0.9\textwidth}
\begin{lstlisting}[language=Python]
def calculate_determinant(matrix):
    if len(matrix) != 3 or len(matrix[0]) != 3:
        raise ValueError('Matrix must be a 3x3 matrix')
    determinant = 0
    for i in range(3):
        submatrix = [[matrix[j][k] for k in range(3) if k != i] for j in range(1, 3)]
        sub_determinant = calculate_determinant(submatrix)
        sign = (-1) ** i
        determinant += sign * matrix[0][i] * sub_determinant
    return determinant if len(matrix) == 2 else round(determinant)
\end{lstlisting}
(Wrong)
        \end{minipage}\\
        \midrule
        CoT &
        \begin{minipage}{0.9\textwidth}
\begin{lstlisting}[language=Python]
def determinant_3x3(matrix):
    if len(matrix) != 3 or not all((len(row) == 3 for row in matrix)):
        raise ValueError('Input matrix must be a 3x3 matrix')
    determinant = 0
    for i in range(3):
        determinant += matrix[0][i] * matrix[1][(i + 1) % 3] * matrix[2][(i + 2) % 3]
        determinant -= matrix[0][i] * matrix[1][(i + 2) % 3] * matrix[2][(i + 1) % 3]
    return determinant
\end{lstlisting}
(Correct)
        \end{minipage}\\
        \midrule
        Ans-Repair &
        \begin{minipage}{0.9\textwidth}
\begin{lstlisting}[language=Python]
def calculate_determinant(matrix):
    if len(matrix) != 3 or len(matrix[0]) != 3:
        raise ValueError('Matrix must be a 3x3 matrix')
    determinant = 0
    for i in range(3):
        submatrix = [[matrix[j][k] for k in range(3) if k != 0] for j in range(3) if j != i]
        sub_determinant = calculate_determinant(submatrix)
        sign = (-1) ** i
        determinant += sign * matrix[0][i] * sub_determinant
    return determinant if len(matrix) == 2 else round(determinant)
\end{lstlisting}
(Wrong)
        \end{minipage}\\
        \midrule
        AMR-Evol &
        \begin{minipage}{0.9\textwidth}
\begin{lstlisting}[language=Python]
def determinant_3x3(matrix: list) -> int:
    det = 0
    for i in range(3):
        det += (-1) ** i * matrix[0][i] * determinant_2x2(submatrix_2x2(matrix, 0, i))
    return det

def determinant_2x2(matrix: list) -> int:
    return matrix[0][0] * matrix[1][1] - matrix[0][1] * matrix[1][0]

def submatrix_2x2(matrix: list, i: int, j: int) -> list:
    submatrix = [row[:j] + row[j + 1:] for row in matrix[:i] + matrix[i + 1:]]
    return submatrix
\end{lstlisting}
(Correct)
        \end{minipage}\\
        \bottomrule
    \end{tabular}
    \caption{Examples of different response distillation methods.}
    \label{tab:example1}
\end{table*}


\begin{table*}
    \small
    \centering
    \begin{tabular}{p{0.05\textwidth}|p{0.9\textwidth}}
        \toprule
        Ins. & Develop a algorithm to simulate multiple rounds of a custom coin flip game with distinct scoring mechanisms for each outcome. The algorithm should efficiently manage a substantial number of players and monitor their scores in real-time, while prioritizing minimal memory usage and computational complexity.\\
        \midrule
        Direct &
        \begin{minipage}{0.9\textwidth}
\begin{lstlisting}[language=Python]
import random
def simulate_round():
    return random.choice(['heads', 'tails'])
def simulate_game(num_players, num_rounds):
    players = {i: 0 for i in range(1, num_players + 1)}
    for _ in range(num_rounds):
        outcome = simulate_round()
        for player in players:
            players[player] = scoring_mechanisms[outcome](players[player])
        print('Scores after round', _ + 1, ':', players)
    return players
\end{lstlisting}
        \end{minipage}\\
        \midrule
        CoT &
        \begin{minipage}{0.9\textwidth}
\begin{lstlisting}[language=Python]
import random
def coin_flip_game(players):
    scores = {}
    for player in players:
        scores[player] = 0
    rounds = 10
    for _ in range(rounds):
        for player in players:
            outcome = random.choice(['heads', 'tails'])
            if outcome == 'heads':
                scores[player] += heads_score
            else:
                scores[player] += tails_score
    return scores
\end{lstlisting}
        \end{minipage}\\
        \midrule
        Ans-Repair &
        \begin{minipage}{0.9\textwidth}
\begin{lstlisting}[language=Python]
import random
def simulate_round():
    return random.choice(['heads', 'tails'])

def simulate_game(num_players, num_rounds):
    players = {i: 0 for i in range(1, num_players + 1)}
    scoring_mechanisms = {'heads': lambda x: x + 1, 'tails': lambda x: max(0, x - 1)}
    for _ in range(num_rounds):
        outcome = simulate_round()
        for player in players:
            players[player] = scoring_mechanisms[outcome](players[player])
        print('Scores after round', _ + 1, ':', players)
    return players
\end{lstlisting}
        \end{minipage}\\
        \midrule
        AMR-Evol &
        \begin{minipage}{0.9\textwidth}
\begin{lstlisting}[language=Python]
import random
class CoinFlipGame:
    def __init__(self):
        self.scores = {}

    def coin_flip(self, num_flips=10):
        heads_count = sum((random.randint(0, 1) for _ in range(num_flips)))
        if heads_count < num_flips / 2:
            return 'You lost the game!'
        else:
            return 'You won the game!'

    def update_scores(self, player, amount):
        if player not in self.scores:
            self.scores[player] = 0
        self.scores[player] += amount

    def get_scores(self):
        return self.scores
\end{lstlisting}
        \end{minipage}\\
        \bottomrule
    \end{tabular}
    \caption{Examples of different response distillation methods. Our AMR-Evol leads more suitable response.}
    \label{tab:example2}
\end{table*}


\begin{table*}
    \small
    \centering
    \begin{tabular}{p{0.05\textwidth}|p{0.9\textwidth}}
        \toprule
        Ins. & Craft a program tasked with computing the Fibonacci sequence up to a user-defined boundary, ensuring that every generated number is not only part of the Fibonacci sequence but also a prime number.\\
        \midrule
        Direct &
        \begin{minipage}{0.9\textwidth}
\begin{lstlisting}[language=Python]
def is_prime(n):
    ...

def fibonacci_prime_sequence(boundary):
    fib_sequence = [0, 1]
    while fib_sequence[-1] <= boundary:
        next_fib = fib_sequence[-1] + fib_sequence[-2]
        if is_prime(next_fib):
            fib_sequence.append(next_fib)
        else:
            fib_sequence.append(next_fib + 1)
    return fib_sequence[:-1] # Wrong
\end{lstlisting}
        \end{minipage}\\
        \midrule
        CoT &
        \begin{minipage}{0.9\textwidth}
\begin{lstlisting}[language=Python]
def is_prime(num):
    ...

def fibonacci_prime_sequence(boundary):
    fib_sequence = [1, 1]
    while fib_sequence[-1] + fib_sequence[-2] <= boundary:
        next_fib = fib_sequence[-1] + fib_sequence[-2]
        if is_prime(next_fib):
            fib_sequence.append(next_fib)
        else:
            fib_sequence.append(next_fib)
    return fib_sequence # Wrong
\end{lstlisting}
        \end{minipage}\\
        \midrule
        Ans-Repair &
        \begin{minipage}{0.9\textwidth}
\begin{lstlisting}[language=Python]
def is_prime(num):
    ...

def fibonacci_prime_sequence(boundary):
    def fibonacci(n):
        if n <= 0: return 0
        elif n == 1: return 1
        else: return fibonacci(n - 1) + fibonacci(n - 2)
    fib_sequence, i = [], 0
    while fibonacci(i) <= boundary:
        next_fib = fibonacci(i)
        if is_prime(next_fib):
            fib_sequence.append(next_fib)
        i += 1
    return fib_sequence # Correct
\end{lstlisting}
        \end{minipage}\\
        \midrule
        AMR-Evol &
        \begin{minipage}{0.9\textwidth}
\begin{lstlisting}[language=Python]
from typing import List
def is_prime(num: int) -> bool:
    ...
    
def generate_fibonacci_sequence(boundary: int) -> List[int]:
    fibonacci_sequence = [0, 1]
    while True:
        next_num = fibonacci_sequence[-1] + fibonacci_sequence[-2]
        if next_num <= boundary:
            fibonacci_sequence.append(next_num)
        else:
            break
    return fibonacci_sequence

def find_fibonacci_primes(boundary: int) -> List[int]:
    fibonacci_sequence = generate_fibonacci_sequence(boundary)
    prime_fibonacci_numbers = [num for num in fibonacci_sequence if is_prime(num)]
    return prime_fibonacci_numbers # Correct
\end{lstlisting}
        \end{minipage}\\
        \bottomrule
    \end{tabular}
    \caption{Examples of different response distillation methods. The \texttt{is\_prime} has been omitted to save space.}
    \label{tab:example3}
\end{table*}

\section{Modular Decomposed and Retrieval Examples}~\label{app:mod_example}


\begin{table*}
    \small
    \centering
    \begin{tabular}{p{0.05\textwidth}|p{0.9\textwidth}}
        \toprule
        Ins. & Craft a program tasked with computing the Fibonacci sequence up to a user-defined boundary, ensuring that every generated number is not only part of the Fibonacci sequence but also a prime number.\\
        \midrule
        Direct & See Table~\ref{tab:example1}\\
        \midrule
        MD &
        \begin{minipage}{0.9\textwidth}
\begin{lstlisting}[language=Python]
def validate_matrix(matrix: list) -> None:
    """
    Description:
        Validates if the input matrix is a 3x3 matrix.
    ...
    """

def calculate_minor_matrix(matrix: list, row: int, col: int) -> list:
    """
    Description:
        Calculates the minor matrix by removing the specified row and column from the input matrix.
    ...
    """

def calculate_determinant(matrix: list) -> int:
    """
    Description:
        Calculates the determinant of a 3x3 matrix using Laplace expansion theorem.
    ...
    """
\end{lstlisting}
        \end{minipage}\\
        \midrule
        Recall &
        \begin{minipage}{0.9\textwidth}
\begin{lstlisting}[language=Python]
def search_element(matrix, x): 
    """
    Search for a given element in a sorted matrix.
    ...
    """
    # Start from the top right corner
    i = 0
    j = len(matrix[0]) - 1 
    while ( i < len(matrix) and j >= 0 ): 
        if (matrix[i][j] == x): 
            return True
       
        if (matrix[i][j] > x): 
            j -= 1
        else: 
            i += 1
    return False

def Submatrix(A: list, i: int, j: int) -> list:
    """
    Get the submatrix of A by removing the i-th row and j-th column.
    ...
    """
    return [row[:j] + row[j+1:] for row in (A[:i] + A[i+1:])]

def Determinant(A: list) -> int:
    """
    Calculate the determinant of the provided matrix A.
    ...
    """
    if len(A) == 1: 
        return A[0][0] 
    if len(A) == 2: 
        return A[0][0]*A[1][1] - A[0][1]*A[1][0] 
    det = 0
    for j in range(len(A)): 
        det += (-1) ** j * A[0][j] * Determinant(Submatrix(A, 0, j))
    return det
\end{lstlisting}
        \end{minipage}\\
        \bottomrule
    \end{tabular}
    \caption{Examples of the modular decomposed (MD) functions and the retrieved top-1 (Recall) functions. We omit some function descriptions to save space.}
    \label{tab:MD_example1}
\end{table*}


\begin{table*}
    \small
    \centering
    \begin{tabular}{p{0.05\textwidth}|p{0.9\textwidth}}
        \toprule
        Ins. & Develop a algorithm to simulate multiple rounds of a custom coin flip game with distinct scoring mechanisms for each outcome. The algorithm should efficiently manage a substantial number of players and monitor their scores in real-time, while prioritizing minimal memory usage and computational complexity.\\
        \midrule
        Direct & See Table~\ref{tab:example2}\\
        \midrule
        MD &
        \begin{minipage}{0.9\textwidth}
\begin{lstlisting}[language=Python]
def simulate_coin_flip() -> str:
    """
    Description:
        Simulates a single coin flip and returns the outcome.
    ...
    """

def update_player_scores(players: dict, outcome: str, scoring_mechanisms: dict) -> None:
    """
    Description:
        Updates the scores of all players based on the outcome of the coin flip.
    ...
    """

def simulate_multiple_rounds(num_players: int, num_rounds: int) -> dict:
    """
    Description:
        Simulates multiple rounds of the game for a given number of players.
    ...
    """
\end{lstlisting}
        \end{minipage}\\
        \midrule
        Recall &
        \begin{minipage}{0.9\textwidth}
\begin{lstlisting}[language=Python]
import random
def coin_flip():
    """Simulate a game of coin flip by flipping a coin 10 times and determining the outcome based on the number of heads. ..."""
    result = 0
    for x in range(10):
        n = random.randint(0, 1)
        if n == 0:
            result += 1      
    if result < 5:
        return "You lost the game!"
    else:
        return "You won the game!"

def score_transactions(transactions):
    """Calculate the total amount of transactions for each sender and store the scores in a dictionary. ..."""
    scores = {}
    for transaction in transactions:
        if transaction['sender'] not in scores:
            scores[transaction['sender']] = 0
        scores[transaction['sender']] += transaction['amount']
    return scores

def determine_winner(scores: list) -> str:
    """Determine the winner of a match based on the scores provided. ..."""
    team_names = [item[0] for item in scores]
    point_diffs = [abs(item[1] - item[2]) for item in scores]
    max_point_diff_idx = point_diffs.index(max(point_diffs))
    if scores[max_point_diff_idx][1] > scores[max_point_diff_idx][2]:
        return team_names[max_point_diff_idx]
    else: return 'Draw'
\end{lstlisting}
        \end{minipage}\\
        \bottomrule
    \end{tabular}
    \caption{Examples of the modular decomposed (MD) functions and the retrieved top-1 (Recall) functions. We omit some function descriptions to save space.}
    \label{tab:MD_example2}
\end{table*}


\begin{table*}
    \small
    \centering
    \begin{tabular}{p{0.05\textwidth}|p{0.9\textwidth}}
        \toprule
        Ins. & Craft a program tasked with computing the Fibonacci sequence up to a user-defined boundary, ensuring that every generated number is not only part of the Fibonacci sequence but also a prime number.\\
        \midrule
        Direct & See Table~\ref{tab:example3}\\
        \midrule
        MD &
        \begin{minipage}{0.9\textwidth}
\begin{lstlisting}[language=Python]
def is_prime(n: int) -> bool:
    """
    Description:
        Checks if a number is a prime number.
    Parameters:
        - n (int): The number to be checked for primality.
    Returns:
        - bool: True if the number is prime, False otherwise.
    """

def fibonacci_prime_sequence(boundary: int) -> List[int]:
    """
    Description:
        Generates a Fibonacci sequence of prime numbers up to a user-defined boundary.
    Parameters:
        - boundary (int): The upper limit for the Fibonacci sequence.
    Returns:
        - List[int]: A list of prime numbers within the Fibonacci sequence up to the boundary.
    """
\end{lstlisting}
        \end{minipage}\\
        \midrule
        Recall &
        \begin{minipage}{0.9\textwidth}
\begin{lstlisting}[language=Python]
def is_prime(num: int) -> bool:
    """
    Check if the provided number is a prime number.
    Args:
        num (int): The number to check for primality.
    Returns:
        bool: True if the number is prime, False otherwise.
    """
    if num <= 1: 
        return False
  
    for i in range(2, num):
        if (num % i) == 0:
            return False
    return True

def prime_sequence(start: int, end: int) -> List[int]:
    """
    Generate a list of prime numbers within the specified range from start to end (inclusive).
    Args:
        start (int): The starting value of the range.
        end (int): The ending value of the range.
    Returns:
        List[int]: A list of prime numbers within the specified range.
    """
    prime_list = []
    for num in range(start, end + 1):
        if num > 1:    
            for i in range(2, num):
                if (num % i) == 0:
                    break
            else:                                         
                prime_list.append(num)
    return prime_list
\end{lstlisting}
        \end{minipage}\\
        \bottomrule
    \end{tabular}
    \caption{Examples of the modular decomposed (MD) functions and the retrieved top-1 (Recall) functions.}
    \label{tab:MD_example3}
\end{table*}

Table~\ref{tab:MD_example1}~\ref{tab:MD_example2}~\ref{tab:MD_example3} showcase the modular decomposed (MD) and retrieved top-1 (Recall) examples.

\section{Comparing with Open Code LLMs}~\label{app:open}

To compare with other Open Code LLMs, we integrate our AMR-Evol framework with Code Evol-Instruct to continually expand our SFT dataset. We also employ the same data decontamination method to prevent data leakage. We have generated approximately 50k training samples. Subsequently, we fine-tuned our models using settings similar to those detailed in Appendix~\ref{app:impl}. Given the larger volume of data, we opted to increase the number of training steps to 400.

To obtain a relative fair comparison, we only include the open code LLMs which are trained with a similar scale of SFT data and employ the same base models as ours, including MagiCoder-DS/CL, WaveCoder-DS, and WizardCoder-CL. We also compare against official instruction-based models, namely DeepSeekCoder-Instruct and CodeLlama-Instrut. However, these official models are trained with more than 20 times data than ours, which lead to unfair comparison. We only want to showcase the performance gaps.

Models with a higher parameter count have been excluded from our comparison, such as DeepSeekCoder-Instruct-33B, WizardCoder-33B-v1.1, Codestral-22B-v0.1,\footnote{\url{https://huggingface.co/mistralai/Codestral-22B-v0.1}}, CodeLlama-Instruct-34B, and Starcoder2-15b-Instruct.\footnote{\url{https://huggingface.co/bigcode/starcoder2-15b-instruct-v0.1}} These models considerably exceed the size of our own, rendering a direct comparison unfair. Additionally, models that primarily derive their learning from GPT4 are excluded, including MagiCoder-S-DS, WaveCoder-DS-Ultra, and OpenCodeInterpreter~\cite{DBLP:journals/corr/abs-2402-14658}. As our teacher model is based on GPT-3.5, a direct comparison with these GPT4-based models would not be equitable. Non-academic models, such as CodeQwen~\cite{qwen}, are also excluded since the methods behind their construction are not disclosed.

In Table~\ref{tab:compare_oss}, all models employ greedy decoding to generate answers for each question. To present additional results and align with some previous studies~\cite{humeval,wizardcoder}, we also display results obtained through sampling in Table~\ref{tab:compare_sampling}. The temperature is set to 0.2, and the number of samples is fixed at 200. Following the method of prior work~\cite{humeval}, we calculate the pass@1 and pass@10 scores. It is also evident that our models outperform the baseline models.

\section{Data Synthesis Cost Trade-off}~\label{app:cost}
Differing from direct distillation, our framework necessitates multi-stage response distillation, which increases the cost of using the API of the teacher model (around 4 times). However, Table~\ref{tab:ds_compare} and~\ref{tab:cl_compare} showcase that our method can outperformance the direct distillation over all tasks and different student models. In addition, we adopt the \texttt{gpt-3.5-turbo-1106} as our teacher model, whose API price is low. Therefore, we conclude that the benefits in performance outweigh the incremental costs incurred.

\section{Adopting Open-Source LLMs as Teachers}\label{app:open_teacher}

While our work primarily focuses on distilling the code generation ability from closed-source models, we also include an additional experiment using the open-source model, Llama-3-70B-Instruct, as our teacher model. Table~\ref{tab:open_teacher} shows that our method is also effective when using the open-source model as the teacher.

\section{Broader Impact}

Our research presents a novel framework for transferring code knowledge from closed-source LLMs to open-source LLMs. This framework is designed to generate code responses for various coding instructions during the data synthesis process. While our approach has been shown to improve response quality, as illustrated in Figure~\ref{fig:human}, it does not guarantee absolute correctness. Consequently, data generated through our method may still contain errors. It is essential to filter out these erroneous samples before deploying our approach in real-world applications to mitigate the risk of misuse.

\section{Manual Evaluation}

\begin{figure*}
    \centering
    \includegraphics[width=0.7\textwidth]{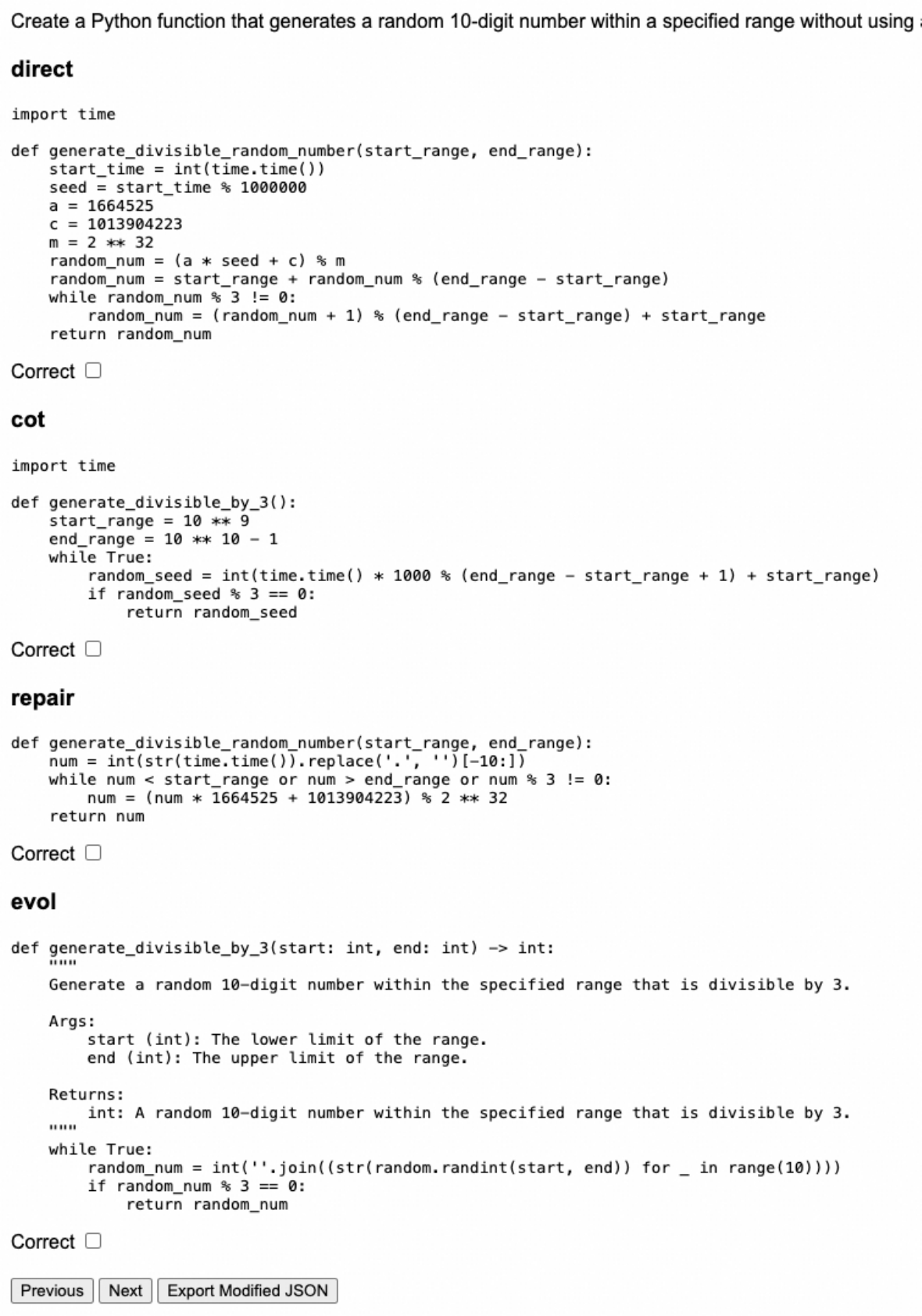}
    \caption{Screenshot of the interface for the human annotators to annotate whether the responses are suitable or not.}
    \label{fig:screenshot}
\end{figure*}

In Figure~\ref{fig:screenshot}, we present the interface used by human annotators to determine whether a given response is an appropriate answer for the coding tasks under evaluation, as shown in Figure~\ref{fig:human}. The annotators are the authors of this paper, possessing expertise in programming.

\section{Use Of AI Assistants}

The AI assistant, GPT4-Turbo, is used solely for refining the writing of our paper.

\end{document}